\documentclass[final]{ecai}

\usepackage{complexity}
\usepackage{bm}         
\usepackage{amssymb}
\usepackage{amsthm}
\usepackage{amsmath}
\usepackage{enumitem}
\usepackage{centernot}
\usepackage{complexity}
\usepackage{mathrsfs}   
\usepackage{tabularx}   
\usepackage{multirow}   
\usepackage{array}      
\usepackage{booktabs}   
\usepackage{colortbl}   
\usepackage{xargs}      
\usepackage{tikz}
\usepackage{tikz-cd}
\usepackage{url}
\usepackage{makecell}   
\usepackage{float}
\usepackage{etoc}
\usepackage[ruled,noline,linesnumbered,noend]{algorithm2e} 
\usepackage{lipsum}
\usepackage{float}
\usepackage{adjustbox}
\usepackage{placeins}
\usepackage{listings}
\usepackage{cleveref}
\usepackage{threeparttable}
\usepackage{minitoc}    

\usetikzlibrary{calc,positioning}

\newtheorem{theorem}{Theorem}[section]

\theoremstyle{definition}

\theoremstyle{definition}
\newtheorem{definition}[theorem]{Definition}

\SetInd{0.5em}{0em}

\newcolumntype{Y}{>{\raggedleft\arraybackslash}X}

\newcommand{\colorofcell}{orange}
\newcommand{\first}[1]{\cellcolor{\colorofcell!45}{#1}}
\newcommand{\second}[1]{\cellcolor{\colorofcell!30}{{#1}}}
\newcommand{\third}[1]{\cellcolor{\colorofcell!15}{{#1}}}

\newcommand{\zerocell}[1]{-}

\definecolor{caribbeangreen}{rgb}{0.0, 0.8, 0.6}
\definecolor{brilliantlavender}{rgb}{0.96, 0.73, 1.0}
\definecolor{amethyst}{rgb}{0.6, 0.4, 0.8}
\definecolor{ao(english)}{rgb}{0.0, 0.5, 0.0}
\definecolor{arylideyellow}{rgb}{0.91, 0.84, 0.42}
\definecolor{asparagus}{rgb}{0.53, 0.66, 0.42}
\definecolor{aquamarine}{rgb}{0.5, 1.0, 0.83}
\definecolor{babyblue}{rgb}{0.54, 0.81, 0.94}
\definecolor{fwtchanged}{rgb}{0.3, 0.3, 0.7}
\definecolor{rosewood}{rgb}{0.4, 0.0, 0.04}
\definecolor{oldmauve}{rgb}{0.4, 0.19, 0.28}
\definecolor{myrtle}{rgb}{0.13, 0.26, 0.12}
\definecolor{magenta(dye)}{rgb}{0.79, 0.08, 0.48}

\definecolor{plta}{rgb}{0.12, 0.47, 0.71}
\definecolor{pltb}{rgb}{   1, 0.5, 0.05}
\definecolor{pltc}{rgb}{0.17, 0.63, 0.17}
\definecolor{pltd}{rgb}{0.84, 0.15, 0.16}

\newcommand{\sssection}[1]{\paragraph{#1}}

\newcommand{\colourdc}{blue}
  
\newcommand{\todocustom}[3]{\todo[linecolor=#2,backgroundcolor=#2!25,bordercolor=#2,size=\tiny,#3]{#1}}  
\newcommandx{\nbdc}[2][1=]{\todocustom{#2}{\colourdc}{#1}}

\def\N{\mathbb{N}}
\def\R{\mathbb{R}}

\renewcommand{\phi}{\varphi}

\def\la{\leftarrow}

\newcommand{\abs}[1]{\left| #1 \right|}

\newcommand{\gen}[1]{\left< #1 \right>}

\newcommand{\set}[1]{\left\{ #1 \right\}}
\newcommand{\seta}[1]{\{ #1 \}}

\newcommand{\mseta}[1]{\{ \!\! \{ #1 \} \!\! \}}

\newcommand{\lr}[1]{\left( #1 \right)}

\newcommand{\range}[1]{\left[\!\left[ #1 \right]\!\right]}

\newcommand{\problem}{\mathbf{P}}
\newcommand{\objects}{\mathcal{O}}
\newcommand{\predicates}{\mathcal{P}}

\newcommand{\schemata}{\mathcal{A}}
\newcommand{\goal}{\mathcal{G}}
\newcommand{\objs}{\mathbf{o}}

\newcommand{\arity}{\mathrm{ar}}

\newcommand{\hash}{\textsc{hash}}
\newcommand{\graphFont}[1]{\mathbf{#1}}
\newcommand{\neighbour}{\graphFont{N}}
\newcommand{\featCat}{\graphFont{F}}

\newcommand{\featEdge}{\graphFont{L}}
\newcommand{\graph}{\graphFont{G}}
\newcommand{\nodes}{\graphFont{V}}
\newcommand{\edges}{\graphFont{E}}

\newcommand{\nodeCat}{\Sigma_{\text{V}}}
\newcommand{\edgeCat}{\Sigma_{\text{E}}}

\newcommand{\colours}{\graphFont{C}}
\newcommand{\countCol}{\textsc{count}}

\newcommand{\Mst}{\graphFont{M}}

\newcommand{\nilgConstSize}{\scriptsize}
\newcommand{\nilgFont}[1]{\texttt{#1}}
\newcommand{\objt}{\nilgFont{object}}
\newcommand{\object}{{\nilgConstSize\nilgFont{object}}}
\newcommand{\pred}{\nilgFont{pred}}

\newcommand{\apg}{{\nilgConstSize\nilgFont{ag}}}
\newcommand{\upg}{{\nilgConstSize\nilgFont{ug}}}
\newcommand{\apv}{{\nilgConstSize\nilgFont{ap}}}

\newcommand{\domainFont}[1]{\textsc{#1}}
\newcommand{\bl}{\domainFont{Bl}}
\newcommand{\ch}{\domainFont{Ch}}
\newcommand{\fe}{\domainFont{Fe}}
\newcommand{\fl}{\domainFont{Fl}}
\newcommand{\mi}{\domainFont{Mi}}
\newcommand{\ro}{\domainFont{Ro}}
\newcommand{\sa}{\domainFont{Sa}}
\newcommand{\so}{\domainFont{So}}
\renewcommand{\sp}{\domainFont{Sp}}
\newcommand{\tr}{\domainFont{Tr}}

\newcommand{\wl}{\texttt{WL}}
\newcommand{\iwl}{\texttt{iWL}}
\newcommand{\niwl}{\texttt{niWL}}
\newcommand{\lwl}{\texttt{2-LWL}}
\newcommand{\kwl}{\texttt{2-WL}}

\newcommand{\none}{\texttt{none}}
\newcommand{\collapselayeryf}{\texttt{i-mf}}
\newcommand{\imf}{\collapselayeryf}

\newcommand{\multiset}{\texttt{mset}}
\newcommand{\sset}{\texttt{set}}

\newcommand{\ppartial}{\texttt{part}}
\newcommand{\complete}{\texttt{cmpl}}

\newcommand{\lasso}{\texttt{Lasso}}
\newcommand{\gpr}{\texttt{GPR}}
\newcommand{\svr}{\texttt{SVR}}
\newcommand{\ranklp}{\texttt{rkLP}}
\newcommand{\rankgpc}{\texttt{rkGPC}}
\newcommand{\ranksvm}{\texttt{rkSVM}}

\begin{document}


\begin{frontmatter}

  \paperid{1052}

  \title{
    Weisfeiler-Leman Features for Planning:\\A 1,000,000 Sample Size Hyperparameter Study
  }

  \author{\fnms{Dillon Z.}~\snm{Chen}}
  \address{LAAS-CNRS, University of Toulouse, France}


  \begin{abstract}
    Weisfeiler-Leman Features (WLFs) are a recently introduced classical machine learning tool for learning to plan and search.
    They have been shown to be both theoretically and empirically superior to existing deep learning approaches for learning value functions for search in symbolic planning.
    In this paper, we introduce new WLF hyperparameters and study their various tradeoffs and effects.
    We utilise the efficiency of WLFs and run planning experiments on single core CPUs with a sample size of 1,000,000 to understand the effect of hyperparameters on training and planning.
    Our experimental analysis show that there is a robust and best set of hyperparameters for WLFs across the tested planning domains.
    We find that the best WLF hyperparameters for learning heuristic functions minimise execution time rather than maximise model expressivity.
    We further statistically analyse and observe no significant correlation between training and planning metrics.
  \end{abstract}
\end{frontmatter}



\section{Introduction}
Learning to plan has gained significant interest in recent years due to the advancements of machine learning approaches across various fields, and the desire to construct autonomous systems that can generalise in long horizon decision making problems.
An aim of learning to plan involves designing automated, domain-independent algorithms for learning domain knowledge in an inductive manner from small training problems that \emph{generalise} and scales up planning to problems of very large sizes~\cite{khardon.ai1999,martin.geffner.ai2004,gretton.thiebaux.uai2004,yoon.etal.jmlr2008,toyer.etal.aaai2018,toyer.etal.jair2020,dong.etal.iclr2019,shen.etal.icaps2020,karia.srivastava.aaai2021,chen.etal.aaai2024,hao.etal.ijcai2024,wang.trevizan.icaps2025}.
Indeed, real-world planning problems do not exhibit much training data for autonomous systems to learn from, meaning that the development of algorithms that can generalise from small training problems is indispensable.

A recently introduced approach involves learning heuristic functions using Weisfeiler-Leman Features (WLFs) automatically extracted from graph representations of planning tasks~\cite{chen.etal.icaps2024}.
The approach involves (1) transforming planning tasks into graphs, and (2) embedding such graphs into feature vectors via the Weisfeiler-Leman (WL) algorithm.
It yields cheap to learn heuristic functions that perform favourably compared to both traditional domain-independent heuristics and learned neural heuristics for planning.
Its performance can be attributed to its faster evaluation speed and greater expressive power of the WL algorithm compared to existing learning for planning work that use neural networks.

In this work, we introduce various extensions and hyperparameters of WLFs with various tradeoffs between model expressivity, generalisation, and execution speed.
Furthermore, we perform large scale experiments resulting in over 1,000,000 planning runs in order to rigorously understand the empirical effects of various hyperparameter settings on (1) training, (2) planning, and (3) the correlation between training and planning metrics.
Our findings are positive and provide us a robust, best set of go-to hyperparameters for generating WLFs for planning.
We observe that the best WLF hyperparameters for learning heuristic functions aim to minimise model size and execution time rather than maximise model expressivity.
Furthermore, we identify that there is no statistically significant, strong correlation between various training metrics and planning performance of WLF models for heuristic search.

\section{Related Work}
The field of learning to plan has been tackled by various different approaches.
In this section we cover more recent approaches for learning to plan and how our approach differs.
We refer to the survey by~\citet{celorrio.etal.ker2012} for earlier works in the field.

\paragraph{Deep and Reinforcement Learning}
It is no surprise that there is a large body of recent work employing deep or reinforcement learning approaches for learning to plan given their progress in various fields.
\citet{toyer.etal.aaai2018,toyer.etal.jair2020} and \citet{dong.etal.iclr2019} introduced the first works employing deep learning for symbolic planning via Graph Neural Networks (GNNs) for learning policies that can generalise to instances of arbitrary size.
Later works employed deep learning architectures to learn heuristics to guide search in a domain-dependent~\cite{groshev.etal.icaps2018,karia.srivastava.aaai2021} and domain-independent~\cite{shen.etal.icaps2020,chen.etal.aaai2024} heuristics.
Expressivity limits of GNNs~\cite{morris.etal.aaai2019,xu.etal.iclr2019,barcelo.etal.iclr2020} were exploited to show that deep learning approaches cannot learn optimal domain knowledge for planning~\cite{staahlberg.etal.icaps2022,chen.etal.aaai2024}.
Reinforcement learning has also been used for learning heuristics~\cite{micheli.valentini.aaai2021,gehring.etal.icaps2022} and policies~\cite{staahlberg.etal.kr2023}, and transformers for learning policies~\cite{rossetti.etal.icaps2024}.

\paragraph{Large Language Models}
Large Language Models have been used to perform one-shot planning via prompting with low success~\cite{valmeekam.etal.neurips2023,valmeekam.etal.c2024}.
They have also been used to perform heuristic evaluation during search but this has been shown again to be inefficient and incomplete~\cite{katz.etal.neurips2024}.
However, they have shown preliminary success in certain domains for \emph{generating} solvers~\cite{silver.etal.aaai2024} or heuristic functions~\cite{tuisov.etal.c2025,correa.etal.c2025} for search as code.

\paragraph{Generalised Planning}
Generalised Planning (GP) refers to a class of approaches which learn interpretable and symbolic general policies that can loop~\cite{levesque.ijcai2005,srivastava.etal.aaai2008,srivastava.etal.ai2011,hu.giacomo.ijcai2011}.
Works in GP learn such policies by generating nondeterministic abstractions of infinite sets of planning instances~\cite{srivastava.etal.aaai2008,illanes.mcilraith.aaai2019,bonet.etal.aaai2019,bonet.geffner.jair2020,cui.etal.ijcai2023} for which a solution is a solution to the ground set of instances.
Abstractions and policies can be synthesised and learned from features~\cite{bonet.etal.aaai2019,frances.etal.ijcai2019} or found via search~\cite{segoviaaguas.etal.icaps2021,segoviaaguas.etal.socs2022,yang.etal.ijcai2022,lei.etal.socs2023,segoviaaguas.etal.ai2024}.

Our work differs from deep learning and large language model approaches as we instead employ \emph{classical} and \emph{statistical} learning approaches which are more efficient in terms of learning and evaluation than deep approaches and have the additional benefit of being interpretable.
We also differ from \emph{policy} approaches in GP that are often complete only under certain assumptions on the planning domain, whereas by employing \emph{search} our approach maintains completeness over planning problems with finite state spaces.
Lastly, our experiments are extensive in terms of scale to determine the major factors contributing to the performance of our approach.

\section{Background}
\label{sec:background}
This section provides the formal definitions of planning tasks and graphs required for understanding the Weisfeiler-Leman Features (WLF)~\cite{chen.etal.icaps2024} used for learning, planning and search.
For the sake of brevity, we focus our attention on classical, lifted planning tasks.
However, we note that WLFs can also handle numeric planning tasks~\cite{chen.thiebaux.neurips2024}.
Furthermore, they are state-centric meaning that they are agnostic to action effects and hence can be extended to handle probabilistic tasks~\cite{zhang.2024}.
Let $\range{n}$ denote the set of integers $\set{1,\ldots,n}$.

\paragraph{Planning Task}
A classical planning task is a deterministic state transition model~\cite{geffner.bonet.2013} given by a tuple $\problem = \gen{S, A, s_0, G}$ where $S$ is a set of states, $A$ a set of actions, $s_0 \in S$ an initial state, and $G \subseteq S$ a set of goal states.
Each action $a \in A$ is a function $a: S \rightarrow S \cup \set{\bot}$ where $a(s) = \bot$ if $a$ is not applicable in $s$, and $a(s) \in S$ is the successor state when $a$ is applied to $s$.
A solution for a planning task is a plan: a sequence of actions $\pi = a_1, \ldots, a_n$ where $s_i = a_i(s_{i-1}) \not= \bot$ for $i \in \range{n}$ and $s_n \in G$.
A state $s$ in a planning task $\problem$ induces a new planning task $\problem' = \gen{S, A, s, G}$.
A planning task is solvable if there exists at least one plan.

\paragraph{Lifted Representation}
Planning tasks are often compactly formalised in a lifted representation using predicate logic, such as via PDDL~\cite{ghallab.etal.1998,haslum.etal.2019}.
More specifically, a lifted planning task is a tuple $\problem = \gen{\objects, \predicates, \schemata, s_0, \goal}$, where $\objects$ denotes a set of objects, $\predicates$ a set of predicate symbols, $\schemata$ a set of action schemata, $s_0$ the initial state, and $\goal$ the goal condition.
We define a domain to be a set of lifted planning tasks which share the same set of predicates $\predicates$ and action schemata $\schemata$.
Understanding of the transition system induced by $\schemata$ is not required for the sake of this paper, so we instead focus on the representation of states and the goal condition next.

Each symbol $P \in \predicates$ is associated with an arity $\arity(P)\in\N \cup \set{0}$.
Predicates take the form $P(x_1,\ldots,x_{\arity(P)})$, where the $x_i$s denote their arguments.
Propositions are defined by substituting objects into predicate arguments.
More specifically, given $P \in \predicates$, and a tuple of objects $\objs = \gen{\objs_1,\ldots,\objs_{\arity(P)}}$, we denote $P(\objs)$ as the proposition defined by substituting $\objs$ into arguments of $P$.
A state $s$ is a set of propositions.
The goal condition $\goal$ also consists of a set of propositions, and a state $s$ is a goal state if $s \supseteq \goal$.

\paragraph{Graphs}\label{ssec:graph}
We denote a graph with categorical node features and edge labels by a tuple $\graph = \gen{\nodes, \edges, \featCat, \featEdge}$.
We have that $\nodes$ is a set of nodes, $\edges \subseteq \nodes \times \nodes$ a set of edges, $\featCat:\nodes \to \nodeCat$ the categorical node features, and $\featEdge:\edges \to \edgeCat$ the edge labels, where $\nodeCat$ and $\edgeCat$ are finite sets of symbols.
The neighbourhood of a node $u \in \nodes$ in a graph is defined by $\neighbour(u) = \set{v \in \nodes \mid \gen{u,v} \in \edges}$.
The neighbourhood of a node $u \in \nodes$ with respect to an edge label $\iota$ is defined by $\neighbour_{\iota}(u) = \set{v \in \nodes \mid e=\gen{u,v} \in \edges \land \featEdge(e) = \iota}$.

\section{WL Features for Planning}
\begin{figure}[t]
  \centering
  \newcommand{\segmentWidth}{1.65cm}
  \newcommand{\segmentHeight}{2.75cm}
  \newcommand{\segmentShift}{3.51cm}
  \newcommand{\trainColour}{cyan}
\newcommand{\objdiff}{0.6}
\newcommand{\factdiff}{0.6}
\newcommand{\xsc}{1.4}
\newcommand{\xscb}{1.7}
\newcommand{\ysc}{2.3}
\newcommand{\objcolour}{babyblue!30}
\newcommand{\fmt}[1]{$\texttt{#1}$}
\newcommand{\nilgTikz}{
    \large
    \node[draw, circle, fill=\objcolour] (A) at
    (\ysc,+2*\factdiff*\xscb) {\fmt{A}};
    \node[draw, circle, fill=\objcolour] (B) at
    (\ysc,+0*\factdiff*\xscb) {\fmt{B}};
    \node[draw, circle, fill=\objcolour] (C) at
    (\ysc,-2*\factdiff*\xscb) {\fmt{C}};
    \node[draw, rectangle, fill=caribbeangreen!80] (a) at
    (0,3*\factdiff*\xsc) {\fmt{onTable(B)}};
    \node[draw, rectangle, fill=caribbeangreen!80] (b) at
    (0,1.5*\factdiff*\xsc) {\fmt{on(C,B)}};
    \node[draw, rectangle, fill=yellow!80] (c) at
    (0,0*\factdiff*\xsc) {\fmt{on(B,A)}};
    \node[draw, rectangle, fill=yellow!80] (d) at
    (0,-1.5*\factdiff*\xsc) {\fmt{onTable(C)}};
    \node[draw, rounded corners=1.5, fill=red!30] (e) at
    (0,-3*\factdiff*\xsc) {\fmt{onTable(A)}};

    \path [-,draw=plta] (B.west) edge (a.east);

    \path [-,draw=plta] (C.west) edge (b.east);
    \path [-,draw=pltb] (B.west) edge (b.east);

    \path [-,draw=plta] (B.west) edge (c.east);
    \path [-,draw=pltb] (A.west) edge (c.east);

    \path [-,draw=plta] (C.west) edge (d.east);

    \path [-,draw=plta] (A.west) edge (e.east);

    \node[draw, circle, fill=\objcolour] (A) at
    (\ysc,+2*\factdiff*\xscb) {\fmt{A}};
    \node[draw, circle, fill=\objcolour] (B) at
    (\ysc,+0*\factdiff*\xscb) {\fmt{B}};
    \node[draw, circle, fill=\objcolour] (C) at
    (\ysc,-2*\factdiff*\xscb) {\fmt{C}};
    \node[draw, rectangle, fill=caribbeangreen!80] (a) at
    (0,3*\factdiff*\xsc) {\fmt{onTable(B)}};
    \node[draw, rectangle, fill=caribbeangreen!80] (b) at
    (0,1.5*\factdiff*\xsc) {\fmt{on(C,B)}};
    \node[draw, rectangle, fill=yellow!80] (c) at
    (0,0*\factdiff*\xsc) {\fmt{on(B,A)}};
    \node[draw, rectangle, fill=yellow!80] (d) at
    (0,-1.5*\factdiff*\xsc) {\fmt{onTable(C)}};
    \node[draw, rounded corners=1.5, fill=red!30] (e) at
    (0,-3*\factdiff*\xsc) {\fmt{onTable(A)}};
}
\newcommand{\bwXshift}{0cm}
\newcommand{\bwYshift}{-0.4cm}
\newcommand{\blocksfontsize}{\scriptsize}
\newcommand{\blocksize}{0.25}
\newcommand{\graphScale}{0.375}
\newcommand{\vecSize}{0.1}
\newcommand{\midShift}{0.25cm}
\begin{tikzpicture}
    \tikzset{
        outer sep=0pt,
        model/.style={
                draw=\trainColour,
                rectangle,
                minimum width=1.5cm,
                minimum height=0.5cm,
                font=\scriptsize,
            },
        heuristic/.style={
                font=\scriptsize,
            },
        vectorSquare/.style={
                draw,
                rectangle,
                minimum width=\vecSize,
                minimum height=\vecSize,
                text width=\vecSize,
                text height=\vecSize,
                text depth=0,
            },
    }
    \node[draw=black, rectangle, rounded corners, minimum width=\segmentWidth, minimum height=\segmentHeight]
    (blocks) at (\bwXshift+0.625cm,0) {};
    \begin{scope}[xshift=\bwXshift, yshift= 2.5*\blocksize cm+\bwYshift]
        \node at (0, 3.75*\blocksize) {\tiny$\problem$};
        \draw (1*\blocksize,0) rectangle (2*\blocksize,1*\blocksize) node[midway] {\blocksfontsize$\texttt{A}$};
        \draw (3*\blocksize,0) rectangle (4*\blocksize,1*\blocksize) node[midway] {\blocksfontsize$\texttt{B}$};
        \draw (3*\blocksize,1*\blocksize) rectangle (4*\blocksize,2*\blocksize) node[midway] {\blocksfontsize$\texttt{C}$};
        \draw (2.5*\blocksize,3*\blocksize) node[align=center] {\tiny initial state};

        \draw(-0.25*\blocksize, 0) rectangle (5.25*\blocksize, -0.125*\blocksize) node[midway,below=0.1*\blocksize] {};
    \end{scope}
    \begin{scope}[xshift=\bwXshift, yshift=-2.5*\blocksize cm+\bwYshift]
        \draw (1*\blocksize,0*\blocksize) rectangle (2*\blocksize,1*\blocksize) node[midway] {\blocksfontsize$\texttt{A}$};
        \draw (1*\blocksize,1*\blocksize) rectangle (2*\blocksize,2*\blocksize) node[midway] {\blocksfontsize$\texttt{B}$};
        \draw (3*\blocksize,0*\blocksize) rectangle (4*\blocksize,1*\blocksize) node[midway] {\blocksfontsize$\texttt{C}$};
        \draw (2.5*\blocksize,3*\blocksize) node[align=center] {\tiny goal condition};

        \draw(-0.25*\blocksize, 0) rectangle (5.25*\blocksize, -0.125*\blocksize) node[midway,below=0.1*\blocksize] {};
    \end{scope}
    \node[draw=black, rectangle, rounded corners, minimum width=\segmentWidth, minimum height=\segmentHeight]
    (graph) at (\segmentShift + 0.425cm,0) {};
    \begin{scope}[yshift=-0.1cm, scale=\graphScale, every node/.append style={transform shape}, xshift=1/\graphScale*\segmentShift*1.05]
        \nilgTikz
        \node at (-1.05, 3.4) {\huge$\graph$};
    \end{scope}
    \node[draw=black, rectangle, rounded corners, minimum width=\segmentWidth, minimum height=\segmentHeight]
    (vector) at (2*\segmentShift + 0.12cm + \midShift, 0) {};

    \begin{scope}[yshift=0, xshift=2.025*\segmentShift + \midShift]
        \node at (-0.55,1.15) {\scriptsize$\mathbf{x}$};
        \draw (0,11*\vecSize) rectangle (\vecSize, 12*\vecSize);
        \draw (0,10*\vecSize) rectangle (\vecSize, 11*\vecSize);
        \draw (0,9*\vecSize) rectangle (\vecSize, 10*\vecSize);
        \draw (0,8*\vecSize) rectangle (\vecSize, 9*\vecSize);
        \draw (0,7*\vecSize) rectangle (\vecSize, 8*\vecSize);
        \draw (0,6*\vecSize) rectangle (\vecSize, 7*\vecSize);
        \draw (0,5*\vecSize) rectangle (\vecSize, 6*\vecSize);
        \draw (0,4*\vecSize) rectangle (\vecSize, 5*\vecSize);
        \draw (0,3*\vecSize) rectangle (\vecSize, 4*\vecSize);
        \draw (0,2*\vecSize) rectangle (\vecSize, 3*\vecSize);
        \draw (0,1*\vecSize) rectangle (\vecSize, 2*\vecSize);
        \draw (0,0*\vecSize) rectangle (\vecSize, 1*\vecSize);
        \draw (0,-1*\vecSize) rectangle (\vecSize, 0*\vecSize);
        \draw (0,-2*\vecSize) rectangle (\vecSize, -1*\vecSize);
        \draw (0,-3*\vecSize) rectangle (\vecSize, -2*\vecSize);
        \draw (0,-4*\vecSize) rectangle (\vecSize, -3*\vecSize);
        \draw (0,-5*\vecSize) rectangle (\vecSize, -4*\vecSize);
        \draw (0,-6*\vecSize) rectangle (\vecSize, -5*\vecSize);
        \draw (0,-7*\vecSize) rectangle (\vecSize, -6*\vecSize);
        \draw (0,-8*\vecSize) rectangle (\vecSize, -7*\vecSize);
        \draw (0,-9*\vecSize) rectangle (\vecSize, -8*\vecSize);
        \draw (0,-10*\vecSize) rectangle (\vecSize, -9*\vecSize);
        \draw (0,-11*\vecSize) rectangle (\vecSize, -10*\vecSize);
        \draw (0,-12*\vecSize) rectangle (\vecSize, -11*\vecSize);
    \end{scope}

    \newcommand{\aboveShift}{0}
    \draw[->,black] (blocks) -- (graph)     node[midway,above=\aboveShift,align=center,font=\scriptsize] {Graph\\Representation\\(\Cref{ssec:graph_rep})};
    \draw[->,black] (graph) -- (vector)     node[midway,above=\aboveShift,align=center,font=\scriptsize] {\phantom{p}Feature\phantom{p}\\\phantom{p}Generation\phantom{p}\\(\Cref{ssec:fg})};

\end{tikzpicture}
  \caption{The WL Feature pipeline for a Blocksworld instance (\texttt{clear} propositions omitted).}
  \label{fig:wlplan}
\end{figure}
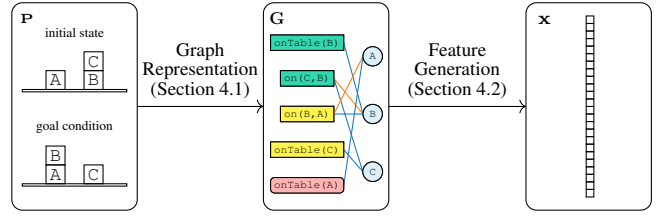

In this section, we describe the procedure for generating vector features for planning tasks without the need for training labels.
\Cref{fig:wlplan} summarises the pipeline which involves converting planning states into graphs, and applying a Weisfeiler-Leman (WL) algorithm to generate the WLFs.

\subsection{Graph Representation}\label{ssec:graph_rep}
The first component of the WLF pipeline involves the transformation of planning tasks into graphs.
Graphs with edge features are viewed as `binary relational structures' in other communities, from which we can derive relational features with various sorts of algorithms.
In this section, we describe the Instance Learning Graph (ILG)~\cite{chen.etal.icaps2024} for representing classical planning tasks.

The middle image in \Cref{fig:wlplan} illustrates a subgraph of the ILG encoding of a simple Blocksworld problem in the left image.
Nodes of the ILG represent the objects coloured in blue, goal conditions, and state information of the planning task, with colours encoding the semantics of nodes.
More specifically, green nodes represent propositions in the state but not part of the goal condition, yellow nodes represent goal conditions that have not yet been achieved in the current state, and red nodes represent goal conditions that have been achieved.
Edges connect objects to propositions that are predicates instantiated with the object, and edge labels encode the location of predicates in which objects are instantiated.
In the image, blue/orange edges connect variable nodes to the object that is instantiated in the first/second argument.
The formal definition is as follows.

\begin{definition}
  \renewcommand{\nilgConstSize}{}
  The Instance Learning Graph (ILG) of a lifted planning task $\problem = \gen{\objects, \predicates, \schemata, s_0, \goal}$ is a graph with categorical node features and edges labels $\graph = \gen{\nodes, \edges, \featCat, \featEdge}$ where
  \begin{itemize}
    \item nodes $\nodes = \objects \cup s_0 \cup \goal$,
    \item edges $\edges = \bigcup_{p=P(\textbf{o}) \in s_0 \cup \goal} \set{\gen{p, \mathbf{o}_i} \mid i \in \range{\arity(P)}}$,
    \item categorical node features $\featCat: \nodes \to \nodeCat$ defined by
      {
        \scriptsize
        \begin{align*}
          \featCat(u) \!=\!
          \begin{cases}
            \objt &\text{if $u \in \objects$} \\
            (\pred(u), \apg) &\text{if $u \in s_0 \cap \goal$} \\
            (\pred(u), \upg) &\text{if $u \in \goal \setminus s_0$} \\
            (\pred(u), \apv) &\text{if $u \in s_0 \setminus \goal$}
          \end{cases}
        \end{align*}
      }
      where $\pred(u)$ denotes the predicate symbol of a proposition $u$.
      We note that $\object$ and $\apg, \upg, \apv$ are constant symbols that are agnostic to the planning domain\footnote{Standing for achieved goal, unachieved goal, and achieved propositional nongoal, respectively.}.
    \item edge labels $\featEdge: \edges \to \N$ defined by $\gen{p, \mathbf{o}_i} \mapsto i$.
  \end{itemize}
\end{definition}

In general, the maximum number of categorical node features in the ILG of any problem for a domain with predicates $\predicates$ is $\abs{\nodeCat} = 1 + 3\abs{\predicates}$ and the number of edge labels is equal to the maximum predicate arity.

\subsection{Feature Generation}\label{ssec:fg}
The second component of the WLF pipeline involves transforming graph representations of planning tasks into feature vectors for use with any downstream task.
The go-to example is learning heuristic functions for heuristic search as we study in this paper.

The algorithms for feature generation of graphs are generally some extension of the colour refinement algorithm, a special case of the general $k$-Weisfeiler-Leman algorithm~\cite{weisfeiler.leman.ni1968,cai.etal.focs1989}.
In this section, we describe the colour refinement, or 1-Weisfeiler-Leman (WL) algorithm, followed by how the algorithm is used for constructing feature vectors from graphs.

\begin{algorithm}[t]
  \caption{WL algorithm}\label{alg:wl}  
\KwIn{A graph $\graph = \gen{\nodes, \edges, \featCat, \featEdge}$, injective $\hash$ function, and number of iterations $L$.}  
\KwOut{Multiset of colours.}  
$c^{0}(v) \la \featCat(v), \forall v \in \nodes$ \label{line:wl:init}\\
\For{$l=1,\ldots,L$ \normalfont{\textbf{do for}} $v \in \nodes$}{ 
    $c^{l}(v) \la 
    \hash
    \lr{c^{l-1}(v), 
    \bigcup_{\iota\in\edgeCat}\mseta{(c^{l-1}(u), \iota) \mid u \in \neighbour_{\iota}(v)}}
    $ \label{line:wl:update}
} 
\Return{$\bigcup_{l=0,\ldots,L}\mseta{c^{l}(v) \mid v \in \nodes}$} \label{line:wl:return}

\end{algorithm}

\paragraph{The WL Algorithm}
The underlying concept of the WL algorithm is to iteratively update node colours based on the colours of their neighbours.
The original WL algorithm was designed for graphs without edge labels.
We present the WL algorithm which can support edge labels~\cite{barcelo.etal.log2022} in \Cref{alg:wl}.
The algorithm's input is a graph with node features and edge labels as described in \Cref{ssec:graph}, alongside a hyperparameter $L$ determining how many WL iterations to perform.
The output of the algorithm is a canonical form for the graph that is invariant to node orderings.

Line 1 initialises node graph colours as their categorical node features.
Lines 2 and 3 iteratively update the colour of each node $v$ in the graph by collecting all its neighbours and the corresponding edge label $(u, \iota)$ into a multiset.
This multiset is then hashed alongside $v$'s current colour with an injective function to produce a new refined colour.
In practice, the injective hash function is built lazily, where every time a new multiset is encountered, it is mapped to a new, unseen hash value.
After $L$ iterations, the multiset of all node colours seen throughout the algorithm is returned.

\paragraph{Embedding graphs}
The WL algorithm has been used to generate features for the WL graph kernel~\cite{shervashidze.etal.jmlr2011}.
Each node colour constitutes a feature, and its value for a graph is the count of the number of nodes that exhibit or has exhibited the colour.
Then given a set of colours $\colours$ known a priori, the WL algorithm can return a fixed sized feature vector of size $\abs{\colours}$ for every input graph.
In a learning for planning pipeline, we collect the colours $\colours$ from a set of training planning tasks, followed by using the colours to embed arbitrary graphs (i.e. converted from either training or testing tasks) into fixed sized feature vectors from such colours.
The steps are formalised as follows.

\begin{enumerate}
  \item We construct $\colours$ from a given set of graph representations of planning tasks $\graph_1, \ldots, \graph_m$ by running the WL algorithm, with the same $\hash$ function and number of iterations $L$, on all of them and then taking the set union of all multiset outputs, i.e. $\colours = \bigcup_{i\in\range{m}} \textsc{WL}(\graph_i)$.
  \item Now suppose we have collected a set of colours and enumerated them by $\colours = \seta{c_1, \ldots, c_{\abs{\colours}}}$.
    Then given a graph $\graph$ and its multiset output from the WL algorithm $\Mst$, we can define an embedding of the graph into Euclidean space by the feature vector
    \begin{align}
      [\countCol(\Mst, c_1), \ldots, \countCol(\Mst, c_{\abs{\colours}})] \in \R^{\abs{\colours}},
      \label{eq:embed}
    \end{align}
    where $\countCol(\Mst, c_i)$ is an integer which counts the occurrence of the colour $c_i$ in $\Mst$.
    We note importantly that any colours returned in $\Mst$ that are not in $\colours$ are defined as \emph{unseen} colours and are entirely ignored in the output.
\end{enumerate}

We can view colours as features, i.e. functions that map planning tasks to real values, by $c_i(\problem) = \countCol(\Mst, c_i)$ where $\Mst$ is the multiset output of WL on $\problem$ encoded to a graph such as via the ILG.

\section{WL Feature Hyperparameters}\label{sec:hyperparameters}
In this section, we introduce the various hyperparameters available in a learning for planning pipeline employing WLFs.
We describe hyperparameters specific to WLFs as internal (\Cref{ssec:internal}) and those that are general to a learning for planning pipeline as external (\Cref{ssec:external}).
The hyperparameters we cover are summarised in \Cref{tab:hyperparameters}.

\newcommand{\hpsection}[1]{
\subsubsection*{#1}}

\begin{table*}[t]
  \newcommand{\ccrule}{\cmidrule{2-6}}
\newcommand{\cmark}{\noindent{\color{blue}{$\uparrow$}}}
\newcommand{\Cmark}{\cmark\cmark}
\newcommand{\nmark}{\noindent{\color{orange}{$\downarrow$}}}
\newcommand{\Nmark}{\nmark\nmark}
\newcommand{\zmark}{\noindent{\color{gray}{--}}}
\begin{tabularx}{\textwidth}{l l p{9.1cm} X X X}
\toprule
& \textbf{Setting} & \textbf{Description} & \textbf{E} & \textbf{G} & \textbf{S} \\
\midrule
\multirow{27}{*}{{Internal}}
& WL Algorithm \\
\ccrule
& \wl{} & The vanilla colour refinement algorithm.
& \zmark & \zmark & \zmark \\
& \iwl{} & WL extended with individualisation but incurs an additional runtime cost as WL is repeated for each node in the graph.
& \cmark & \nmark & \nmark \\
& \niwl{} & Equivalent to \iwl{} except that node features are normalised by the number of node features.
& \cmark & \nmark & \nmark \\
& \lwl{} & A feasible approximation of 2-WL but still incurs the worst case runtime cost which is quadratic in the number of nodes.
& \cmark & \nmark & \nmark \\
& \kwl{} & WL extended to pairs of nodes. Computationally infeasible both runtime and memory-wise for the tested datasets.
& \Cmark & \Nmark & \Nmark \\
\ccrule
& Iterations \\
\ccrule
& \texttt{4}  & An arbitrarily chosen number in the original WLF paper that works well for optimal heuristic estimators.
& \zmark & \zmark & \zmark \\
& Low  ($<4$) & Trades expressivity for improved speed and model size.
& \nmark & \cmark & \cmark \\
& High ($>4$) & Trades speed and generalisation for improved expressivity.
& \cmark & \nmark & \nmark \\
\ccrule
& Feature Pruning \\
\ccrule
& \texttt{none} & No pruning of collected features.
& \zmark & \zmark & \zmark \\
& \texttt{i-mf} & Combination of MaxSAT and frequency counting for pruning.
& \nmark & \cmark & \cmark \\
\ccrule
& Hash Function \\
\ccrule
& \texttt{multiset} & The hash input in the original WL algorithm which corresponds to collecting subtrees of a graph.
& \zmark & \zmark & \zmark \\
& \texttt{set} & Collapses duplicate neighbour colours with the aim of reducing the number of unseen colours outside of training.
& \nmark & \cmark & \cmark \\
\midrule
\multirow{12.5}{*}{{External}}
& State Representation \\
\ccrule
& \ppartial{} & Prunes propositions deemed static or unreachable by Fast Downward before generating WLFs for faster generation.
& \nmark & \zmark & \cmark \\
& \complete{} & Uses all propositions from each state for generating WLFs. Does not lose information but is slower to generate.
& \cmark & \zmark & \nmark \\
\ccrule
& Optimiser \\
\ccrule
& \lasso{} & Linear Regression with L1 prior for predicting optimal heuristics. \\
& \gpr{} & Gaussian Process Regression for predicting optimal heuristics. \\
& \svr{} & Support Vector Regression for predicting optimal heuristics. \\
& \ranklp{} & Linear Programs for predicting ranking heuristics. \\
& \rankgpc{} & Gaussian Process Classification for predicting ranking heuristics. \\
& \ranksvm{} & Support Vector Machines for predicting ranking heuristics. \\
\bottomrule
\end{tabularx}

  \caption{A summary of various WLF hyperparameters and descriptions. Default values of internal features are marked as so.}
  \label{tab:hyperparameters}
\end{table*}

\subsection{Internal Hyperparameters}\label{ssec:internal}
\hpsection{WL Algorithm}
The WL algorithm from \Cref{ssec:fg} is the canonical graph kernel baseline for graph learning tasks due to the theoretical result that it upper bounds distinguishing power of the message passing GNN architecture~\cite{morris.etal.aaai2019,xu.etal.iclr2019}.
It is also an efficient algorithm that runs in low polynomial time in the input graph and considered the first choice to apply to graphs as described in the extensive graph kernel survey by~\citet{kriege.etal.ans2020}.
Nevertheless, the graph learning community has proposed various extensions of the WL algorithm and corresponding GNN architectures that have provably more distinguishing power than the WL algorithm, and yet are still computationally feasible.

\begin{figure}[ht]
  \centering
  \scalebox{0.5}{
    \LARGE
    \trimbox{0.15cm 1.8cm 15.5cm 1.8cm}{
      \begin{tikzpicture}

    \newcommand{\majorSize}{4.cm}
    \newcommand{\majorShift}{4.25cm}
    \newcommand{\minorShift}{2.5cm}
    \newcommand{\nodeShift}{1cm}

    \newcommand{\circSize}{0.5cm}
    \newcommand{\dualSize}{1.5cm}
    \newcommand{\sqrtTwo}{1.41421356237}

    \newcommand{\minorSize}{0.5}
    \newcommand{\minorNudge}{1.12cm}
    \newcommand{\minorEdgeNudge}{0.55cm}

    \tikzset{
        badge/.style={
            rectangle,
            rounded corners,
            draw,
        },
        major/.style={
            badge,
            minimum width=\majorSize,
            minimum height=\majorSize,
        },
        node/.style={
            draw,
            circle,
            inner sep=0,
            minimum size=0.3cm,
        },
        minorNode/.style={
            node,
            minimum size=0.3cm*\minorSize,
        },
        circleNode/.style={
            badge,
            minimum width=\circSize,
            minimum height=\circSize,
        },
        minorCircleNode/.style={
            rectangle,
            draw,
            rounded corners=2pt,
            minimum size=\minorSize*\circSize,
        },
        dottedNode/.style={
            circleNode,
            dashed,
        },
        minorDottedNode/.style={
            minimum size=0.5*\minorSize*\circSize,
            rounded corners=1pt,
            rectangle,
            draw,
            dotted,
        },
        title/.style={
            align=left,
            anchor=north west,
        },
        bottom/.style={
            align=left,
            anchor=south west,
            font=\tiny,
            white,
        }
    }

    \newcommand{\tc}[1]{\tikz[baseline=-0.5ex]\draw[radius=1.5pt,fill=#1] (0,0) circle ;}%
    \renewcommand{\tc}[1]{}

    \newcommand{\lime}{1.9cm}
    \newcommand{\titleCoord}{-\lime, \lime}
    \newcommand{\bottomCoord}{-\lime, -\lime}
    \newcommand{\aCoord}{0, 0}
    \newcommand{\bCoord}{0, \nodeShift}
    \newcommand{\cCoord}{\nodeShift, 0}
    \newcommand{\dCoord}{0, -\nodeShift}
    \newcommand{\eCoord}{-\nodeShift, 0}

    \newcommand{\gcol}{green!20}
    \newcommand{\bcol}{blue!20}
    \newcommand{\ocol}{orange!20}
    \newcommand{\pcol}{purple!20}

    \newcommand{\graphMacro}{
        \node[node,fill=\gcol]  (a) at (\aCoord) {};
        \node[node,fill=\gcol]  (b) at (\bCoord) {};
        \node[node,fill=\bcol]   (c) at (\cCoord) {};
        \node[node,fill=\ocol] (d) at (\dCoord) {};
        \node[node,fill=\pcol] (e) at (\eCoord) {};

        \draw(a) -- (b);
        \draw(a) -- (c);
        \draw(a) -- (d);
        \draw(c) -- (d);
        \draw(b) -- (e);
    }

    \begin{scope}
        \node[major] at (0, 0) {};
        \node[title] at (\titleCoord) {WL};
        \node[bottom] at (\bottomCoord) {$(\tc{\gcol}, \mseta{\tc{\gcol}, \tc{\ocol}, \tc{\bcol}})$};
        \graphMacro{}
        \node[circleNode] at (\aCoord) {};
        \node[dottedNode] at (\bCoord) {};
        \node[dottedNode] at (\cCoord) {};
        \node[dottedNode] at (\dCoord) {};
    \end{scope}

    \begin{scope}[xshift=1*\majorShift]
        \renewcommand{\mseta}[1]{\{ \!\! \{ \! #1 \! \} \!\! \}}
        \node[major] at (0, 0) {};
        \node[title] at (\titleCoord) {2-WL};
        \graphMacro{}
        \node[circleNode, minimum width=\dualSize] at ($(a)!0.5!(c)$) {};
        \node[dottedNode] at (\bCoord) {};
        \node[dottedNode] at (\dCoord) {};
        \node[dottedNode] at (\eCoord) {};
        \node[bottom] at (\bottomCoord) {$(\!\mseta{\tc{\gcol}, \tc{\pcol}},\!\mseta{
        \mseta{\tc{\gcol},\!\tc{\bcol},\!\tc{\gcol}}
        ,\!\mseta{\tc{\gcol},\!\tc{\bcol},\!\tc{\ocol}}
        ,\!\mseta{\tc{\gcol},\!\tc{\bcol},\!\tc{\pcol}}
        }\!)$};
    \end{scope}

    \begin{scope}[xshift=2*\majorShift]
        \node[major] at (0, 0) {};
        \node[title] at (\titleCoord) {2-LWL};
        \graphMacro{}
        \node[circleNode, minimum width=\dualSize] at ($(a)!0.5!(c)$) {};
        \node[dottedNode] at (\bCoord) {};
        \node[dottedNode] at (\dCoord) {};
        \node[bottom] at (\bottomCoord) {$(\!\mseta{\tc{\gcol}, \tc{\pcol}},\!\mseta{
            \mseta{\tc{\gcol},\!\tc{\bcol},\!\tc{\gcol}}
            ,\!\mseta{\tc{\gcol},\!\tc{\bcol},\!\tc{\ocol}}
        }\!)$};
    \end{scope}

    \begin{scope}[xshift=3*\majorShift]
        \node[major] at (0, 0) {};
        \node[title] at (\titleCoord) {iWL};

        \node[minorCircleNode] (center) at (\aCoord) {};
        \coordinate[left=\minorNudge of center] (lcenter) ;
        \coordinate[right=\minorNudge of center] (rcenter) ;
        \coordinate[above=\minorNudge of center] (acenter) ;
        \coordinate[below=\minorNudge of center] (bcenter) ;
        \node[minorCircleNode] at (lcenter) {};
        \node[minorCircleNode] at (rcenter) {};
        \node[minorCircleNode] at (acenter) {};
        \node[minorCircleNode] at (bcenter) {};

        \renewcommand{\nodeShift}{0.45cm}
        \node[minorDottedNode] (center) at (\bCoord) {};
        \coordinate[left=\minorNudge of center] (lcenter) ;
        \coordinate[right=\minorNudge of center] (rcenter) ;
        \coordinate[above=\minorNudge of center] (acenter) ;
        \coordinate[below=\minorNudge of center] (bcenter) ;
        \node[minorDottedNode] at (lcenter) {};
        \node[minorDottedNode] at (rcenter) {};
        \node[minorDottedNode] at (acenter) {};
        \node[minorDottedNode] at (bcenter) {};
        \node[minorDottedNode] (center) at (\cCoord) {};
        \coordinate[left=\minorNudge of center] (lcenter) ;
        \coordinate[right=\minorNudge of center] (rcenter) ;
        \coordinate[above=\minorNudge of center] (acenter) ;
        \coordinate[below=\minorNudge of center] (bcenter) ;
        \node[minorDottedNode] at (lcenter) {};
        \node[minorDottedNode] at (rcenter) {};
        \node[minorDottedNode] at (acenter) {};
        \node[minorDottedNode] at (bcenter) {};
        \node[minorDottedNode] (center) at (\dCoord) {};
        \coordinate[left=\minorNudge of center] (lcenter) ;
        \coordinate[right=\minorNudge of center] (rcenter) ;
        \coordinate[above=\minorNudge of center] (acenter) ;
        \coordinate[below=\minorNudge of center] (bcenter) ;
        \node[minorDottedNode] at (lcenter) {};
        \node[minorDottedNode] at (rcenter) {};
        \node[minorDottedNode] at (acenter) {};
        \node[minorDottedNode] at (bcenter) {};

        \renewcommand{\nodeShift}{1cm}
        \begin{scope}[xshift=3*\majorShift]
            \scalebox{\minorSize}{\graphMacro{}}
            \begin{scope}[xshift=\minorShift]
                \scalebox{\minorSize}{\graphMacro{}}
            \end{scope}
            \begin{scope}[xshift=-\minorShift]
                \scalebox{\minorSize}{\graphMacro{}}
            \end{scope}
            \begin{scope}[yshift=\minorShift]
                \scalebox{\minorSize}{\graphMacro{}}
            \end{scope}
            \begin{scope}[yshift=-\minorShift]
                \scalebox{\minorSize}{\graphMacro{}}
            \end{scope}
        \end{scope}

        \node[minorNode, fill=white, draw=white] (center) at (\aCoord) {};
        \coordinate[left=\minorNudge+\minorEdgeNudge of center] (lcenter) ;
        \coordinate[right=\minorNudge+\minorEdgeNudge of center] (rcenter) ;
        \coordinate[above=\minorNudge+\minorEdgeNudge of center] (acenter) ;
        \coordinate[below=\minorNudge+\minorEdgeNudge of center] (bcenter) ;
        \node[minorNode, fill=white, draw=white] at (lcenter) {};
        \node[minorNode, fill=white, draw=white] at (rcenter) {};
        \node[minorNode, fill=white, draw=white] at (acenter) {};
        \node[minorNode, fill=white, draw=white] at (bcenter) {};
        \node[font=\tiny] at (center) {$\otimes$};
        \node[font=\tiny] at (lcenter) {$\otimes$};
        \node[font=\tiny] at (rcenter) {$\otimes$};
        \node[font=\tiny] at (acenter) {$\otimes$};
        \node[font=\tiny] at (bcenter) {$\otimes$};

    \end{scope}

\end{tikzpicture}
    }
  }
  \quad
  \caption{
    Visualisations of neighbours (dotted) of the center node or node pair (solid) in the WL, 2-WL, 2-LWL and iWL algorithms.
    In iWL, the WL algorithm is run $\abs{\nodes}$ times, where each time a different node is individualised with a special colour.
  }
  \label{fig:wls}
  \label{fig:hierarchy}
\end{figure}
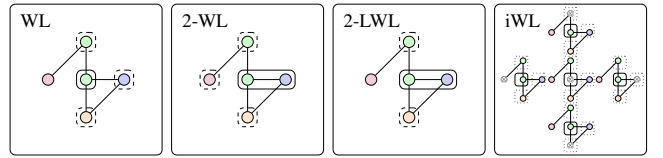

Most notably, it is well known that the $(k+1)$-WL algorithm is strictly more powerful than the $k$-WL algorithm for $k \geq 2$ but its runtime scales exponentially in $k$.
Thus, many extensions of the WL algorithm and corresponding GNNs have been proposed to either approximate or achieve orthogonal expressiveness of higher order WL algorithms~\cite{morris.etal.neurips2020,morris.etal.icml2022,zhao.etal.neurips2022,wang.etal.iclr2023,alvarezgonzalez.etal.tmlr2024}.
Furthermore, graph kernels have been proposed that also handle graphs with continuous node attributes~\cite{chen.thiebaux.neurips2024}.
We have implemented some of these WL extensions alongside completely new graph kernels in the current version of WLPlan which we outline as follows.
\Cref{fig:hierarchy} illustrates the expressivity hierarchy of mentioned WL algorithms.

\newcommand{\algoFontSize}{\small}

\sssection{2-WL}
The $k$-WL algorithms are a suite of incomplete graph isomorphism algorithms which have a one-to-one correspondence to $k$-variable counting logics~\cite{cai.etal.focs1989}.
However, the $k$-WL algorithms scale exponentially in $k$, with the 2-WL algorithm exhausting memory limits on medium sized graph datasets.
We describe and implement the 2-WL algorithm and refer to \cite[Section 5, page 13]{cai.etal.focs1989} and \cite[Section 5, page 4]{grohe.lics2021} for the general $k$-WL algorithm.

The idea of the 2-WL algorithm, outlined in \Cref{alg:2wl}, is to now assign and refine colours to ordered pairs of nodes.
The algorithm begins in Lines 1-3 by assigning all possible ordered pairs of nodes a tuple of the node colours as well as the edge label between them.
If there is no edge between a pair of nodes, a special $\bot$ value is used as the edge label.
Lines 4-5 then iteratively refine the colour of each node pair by defining the neighbours of a pair $(v, u)$ to be a \emph{multiset of sequence of pairs}\footnote{In contrast, the \emph{Oblivious} $k$-WL (cf. \cite[Section 5, page 5]{grohe.lics2021}) defines the neighbours as a \emph{sequence of multiset of pairs} for $k=2$.} $((w, u), (v, w))$ where $w$ ranges over all nodes in the graph.
Then the algorithm applies the colouring function of the current iteration to all node pairs to create a multiset of tuples of colours which are then hashed alongside the current node pair's colour.
Finally, the algorithm returns the multiset of all node pair colours seen throughout the algorithm in Line 6.
To generalise to the $k$-WL algorithm, one replaces node pairs with node $k$-tuples.
\begin{algorithm}[t]
  \algoFontSize
  \caption{2-WL algorithm}\label{alg:2wl}  
\KwIn{A graph $\graph = \gen{\nodes, \edges, \featCat, \featEdge}$, injective $\hash$ function, and number of iterations $L$.}  
\KwOut{Multiset of colours.}  
$e(v, u) \la \featEdge(v, u), \forall (v,u) \in \edges$ \\
$e(v, u) \la \bot, \forall (v, u) \in (\nodes^2) \setminus \edges$ \\
$c^{0}(v, u) \la (\featCat(v), \featCat(u), e(v, u)), \forall (v,u) \in \nodes^2$ \label{line:2wl:init}\\
\For{$j=1,\ldots,L$ \normalfont{\textbf{do for}} $(v,u) \in \nodes^2$}{ 
    $c^{j}(v,u) \la 
    \hash
    \bigl(c^{j-1}(v,u), 
    \{\!\{
    (c^{j-1}(w,u), c^{j-1}(v,w))
    \mid 
    w \in \nodes
    \}\!\}
    \bigr)
    $ \label{line:2wl:update}
} 
\Return{$\bigcup_{j=0,\ldots,L}\mseta{c^{j}(v, u) \mid (v, u) \in \nodes^2}$} \label{line:2wl:return}

\end{algorithm}

\sssection{2-LWL}
The $k$-LWL algorithms~\cite{morris.etal.icdm2017} provide efficient approximations of the $k$-WL algorithms but still have the same worst case computational complexity.
The main approximations made are that node tuples are converted to node sets, reducing the number of possible node tuples to consider per iteration by a constant factor ($n^k \to {{n}\choose{k}}$), and relaxing the definition of neighbours of node tuples.
Now the neighbour for a 2-sets of nodes $\set{v, u}$ in 2-LWL is defined by the set of set of 2-sets $\set{\set{w, u}, \set{v, w}}$ where $w$ now ranges over the union of neighbours of $u$ and $v$, instead of over all nodes.
\Cref{alg:2lwl} outlines the 2-LWL algorithm and \Cref{fig:wls} illustrates the different neighbour definitions of 2-WL and 2-LWL.
\begin{algorithm}[t]
  \algoFontSize
  \newcommand{\nodePair}[1]{\seta{#1}}
\caption{2-LWL algorithm}\label{alg:2lwl}  
\KwIn{
    An undirected graph $\graph = \gen{\nodes, \edges, \featCat, \featEdge}$, injective $\hash$ function, and number of iterations $L$.
    Let $\nodePair{u, v}$ denote a node pair without order or undirected edge.
}  
\KwOut{Multiset of colours.}  
$e\nodePair{v,\!u} \!\la\! \featEdge\nodePair{v,\!u}, \forall \nodePair{v,\!u} \!\in\! \edges; e\nodePair{v,\!u}\!\la\!\bot, \forall \nodePair{v,\!u} \!\in\! {{\nodes}\choose{2}}\!\!\setminus\!\edges\!\!$ \\
$c^{0}\nodePair{v, u} \!\la\! (\featCat(v), \featCat(u), e\nodePair{v, u}), 
\forall \nodePair{v, u} \!\in\! {{\nodes}\choose{2}}$ 
\label{line:2lwl:init}\\
\For{$j=1,\ldots,L$ \normalfont{\textbf{do for}} $\nodePair{v, u} \in {{\nodes}\choose{2}}$}{ 
    $c^{j}\nodePair{v,\!u} \!\la\!
    \hash
    \bigl(c^{j-1}\nodePair{v,\!u}, \!
    \{\!\!\{\!
    \mseta{c^{j-1}\nodePair{w, u}, c^{j-1}\nodePair{v, w}}
    \!\!
    \mid 
    w \in \neighbour(v) \cup \neighbour(u)
    \}\!\!\}
    \bigr)
    $ \label{line:2lwl:update}
} 
\Return{${\displaystyle}_{j=0,\ldots,L}\mseta{c^{j}\nodePair{v,\!u} \mid \nodePair{v,\!u} \in {{\nodes}\choose{2}}}$} \label{line:2lwl:return}

\end{algorithm}

\sssection{iWL}
We introduce an expressive WL algorithm extension inspired by Identity-aware GNNs (ID-GNNs)~\cite{you.etal.aaai2021}, orthogonal to the $k$-WL algorithms.
ID-GNNs run a GNN $\abs{\nodes}$ times on a graph which uses different parameters for embedding updates on a selected, individualised node on each GNN run.
We kernelise the ID-GNN algorithm into what we call the individualised WL algorithm, presented in \Cref{alg:iwl} and \Cref{fig:wls}.

We have a single outer loop iterating over all nodes $w \in \nodes$ in a graph in Line 1, and within each inner loop, all nodes are assigned their initial node colour, except for $w$ which is augmented a special, individualised colour $\otimes$ that is not in $\nodeCat$ of $\featCat$.
The remainder of the algorithm is equivalent to the WL algorithm, except that iWL now returns $\abs{\nodes}$ more colours in the output multiset due to the outer loop.

Given that the number of colours returned by iWL is quadratic in the number of refinable items, we proposed normalising the embeddings of the multisets by dividing $\countCol(\Mst, c)$ by $\abs{\nodes}$.
In the experiments below, we notate this feature generator as niWL.
\begin{algorithm}[t]
  \algoFontSize
  \caption{iWL algorithm}\label{alg:iwl}  
\KwIn{A graph $\graph = \gen{\nodes, \edges, \featCat, \featEdge}$, injective $\hash$ function, and number of iterations $L$.}  
\KwOut{Multiset of colours.}  
\For{$w \in \nodes$}{
    $c^{0}_w(v) \la \featCat(v), \forall v \in \nodes \setminus \set{w}$ ;
    $c^{0}_w(w) \la (\featCat(w), \otimes)$  \label{line:iwl:init}\\
    \For{$j=1,\ldots,L$ \normalfont{\textbf{do for}} $v \in \nodes$}{ 
        $c^{j}_w(v) \la 
        \hash
        \lr{c^{j-1}_w(v), 
        \bigcup_{\iota\in\edgeCat}\mseta{(c^{j-1}_w(u), \iota) \mid u \in \neighbour_{\iota}(v)}}
        $ \label{line:iwl:update}
    } 
}
\Return{$\bigcup_{w \in \nodes}\bigcup_{j=0,\ldots,L}\mseta{c^{j}_w(v) \mid v \in \nodes}$} \label{line:iwl:return}

\end{algorithm}

\hpsection{Iterations}
The number of iterations in a WLF configuration refers to the $L$ parameter in \Cref{alg:wl} and similar WL algorithms.
One can view the number of iterations as analogous to the number of message passing layers in a graph neural network, which determines the receptive field of the network around each graph node.

\hpsection{Feature Pruning}
Feature pruning is a technique for reducing the number of redundant or irrelevant features in a machine learning model in the context of planning.
Different feature pruning approaches trade off soundness, referring the maximal preservation of feature information, and computation arising from faster evaluation and lower memory footprint while training models.
In our evaluation, we experiment with no pruning and the iterative MaxSAT plus frequency pruning (\texttt{i-mf})~\cite{hao.etal.ecai2025}.
More specifically, \texttt{i-mf} prunes features whose evaluations on the training set are equivalent to existing features as seen in previous works (e.g. \cite{martin.geffner.kr2000,bonet.etal.aaai2019,frances.etal.aaai2021,drexler.etal.icaps2022}) but uses a MaxSAT encoding to add the constraint that features are pruned only if no other features that depend on it are pruned. After pruning via MaxSAT, features that appear less than 1\% of the time are also pruned.

\hpsection{Hash Function}
The hash function hyperparameter determines if we remain using the \texttt{multiset} hash input in \Cref{line:wl:update} of the WL algorithm in \Cref{alg:wl} and similar variants, or replace the multiset hash with a \texttt{set} hash.
In both cases however, the \emph{output} still uses a multiset in order to return numeric feature vectors.
The reasoning for using a set hash is generate a smaller number of features at the expense of expressivity to improve runtime and generalisation, and to reduce the number of unseen colours during inference.
For example, consider a training set consisting of star graphs of degree at most 5.
If at inference a new star graph of degree 6 is introduced, the center node becomes an unseen colour after one WL iteration, which prematurely limits its receptive field.
Indeed, \citet{drexler.etal.kr2024} show that for a sample of states on most planning domains that reducing the hash function from multisets to sets does not compromise model expressivity.

\subsection{External Hyperparameters}\label{ssec:external}
\hpsection{State Representation}
The external state representation hyperparameter refers to the proposition which are used for feature generation.
The canonical approach is to use all propositions in the state for WLF generation which we denote as the \texttt{complete} representation.
Alternatively, as performed in the original WLF paper~\cite{chen.etal.icaps2024}, one can generate features for a subset of facts such as those detected as relevant and nonstatic by Fast Downward's grounder~\cite{helmert.ai2009}.
We name this configuration \texttt{partial}.
The reasoning for this is that certain static propositions may be redundant for feature generation, such as the `up' and `down' propositions in the Miconic domain, which are only relevant for action applicability.
Removing such propositions in certain domains significantly speeds up the feature generation runtime and hence increases the number of node expansions to be performed during search.
Conversely, there are cases where relevant static facts could be pruned and thus, this option provides a tradeoff between speed and expressivity, depending on the planning domain.

\hpsection{Optimiser}
The external optimiser hyperparameter refers to the method used to compute parameters of a prediction model using WLFs.
In this paper we focus on linear models for predicting heuristic functions for search.
More specifically, given a set of WLFs $c_1, \ldots, c_n$, we aim to find a set of weights $w_1, \ldots, w_n \in \R$ that gives us the `best' heuristic function $\sum_{i\in\range{n}} w_ic_i(s)$.

The original WLF paper experimented with two optimiser formulations for computing linear models using Support Vector Machines (\svr) and linear kernel Gaussian Process Regression (\gpr), as well as higher order kernel approaches.
However, higher order kernels yielded poorer performance, and similarly for other non-linear predictors such as XGBoost and neural networks from additional informal experiments.
Our hypothesis for these observations is that WL features are already generating sufficiently informative features such that a linear model is best for generalisation and other approaches lead to overfitting to the training label range.

In this paper, we focus on new linear models formulated by Linear Regression with L1 regularisation (\lasso), and additional ranking formulations: the Rank Support Vector Machine (\ranksvm) setup for learning heuristic functions introduced by~\citet{garrett.etal.ijcai2016}, a similar formulation using Gaussian Process Classification (\rankgpc) where one replaces the SVM optimiser with a linear kernel GPC optimiser, and the Linear Programming ranking formulation (\ranklp) introduced by~\citet{chen.thiebaux.neurips2024}.

\section{Experiments}
In this section, we perform a one million sample size experiment to evaluate the effects of the various WLF hyperparameters covered in \Cref{sec:hyperparameters} and summarised in \Cref{tab:hyperparameters} on learning and planning.
The source code, scripts for running the experiments, and appendix for the paper are publicly available in~\cite{chen.2025}.

\paragraph{Benchmarks}
Our benchmarks consist of the 10 domains and the training and testing splits from the International Planning Competition Learning Track 2023~\cite{taitler.etal.am2024}.
More specifically, the domains are Blocksworld (\bl), Childsnack (\ch), Ferry (\fe), Floortile (\fl), Miconic (\mi), Rovers (\ro), Satellite (\sa), Sokoban (\so), Spanner (\sp), and Transport (\tr).
We generate optimal plans from the given training tasks via the Scorpion planner~\cite{seipp.etal.jair2020} with a time limit of 30 minutes and 8GB memory limit for each training task on a cluster with Intel Xeon Platinum 8274 CPUs.
We note however that the median planning time of solved tasks lies in a few seconds for each domain.
Only states and their corresponding $h^*$ values in optimal plan traces were used for non-ranking models ($\lasso, \gpr, \svr$), while both states and their siblings were used for ranking models ($\ranklp, \rankgpc, \ranksvm$).
Each domain contains 90 testing tasks, for a total of 900 testing tasks.
Further benchmark details are provided in \Cref{app:benchmarks}.
We refer to previous work~\cite{chen.etal.icaps2024} for a detailed discussion and comparison of learned WL heuristics with classical planners, as the benchmarks are the same and this work focuses on the effect of hyperparameters on WLF models.

\paragraph{Configurations}
Our experiments involve training various models for learning heuristic functions for search.
The configurations we experiment with are all possible combinations of the 6 hyperparameters discussed in the previous section.
More specifically, the hyperparameter options and their ranges are
\begin{enumerate}[label={(\arabic*)}]
  \item WL Algorithm: $\set{\wl, \iwl, \niwl, \lwl}$,
  \item Iterations: $\set{\texttt{1},\texttt{2},\texttt{3},\texttt{4},\texttt{5},\texttt{6},\texttt{7},\texttt{8}}$,
  \item Feature Pruning: $\set{\none, \collapselayeryf}$,
  \item Hash Function: $\set{\multiset, \sset}$,
  \item State Representation: $\set{\ppartial, \complete}$,
  \item Optimiser: $\{$\lasso, \gpr, \svr, \ranklp, \rankgpc, \ranksvm$\}$.
\end{enumerate}

Model training for all configurations was given a memory limit of 8GB and 5 minutes.
The 2-WL configuration was not included in the results because their models were too memory intensive to train in the given limits.
Learned heuristic functions that use the \complete{} state representation were used with the Powerlifted search engine~\cite{correa.etal.icaps2020}, while those which used the \ppartial{} state representation were used with Version 24.06 of the Fast Downward search engine.
Models using the \ppartial{} state representation were given the FDR input of a planning task after at most 5 minutes of grounding~\cite{helmert.ai2009}.
The reason for using different search engines is that profiling showed that a significant proportion of planning time is spent on heuristic evaluation and not successor generation.
Due to the large volume of experiments and resource constraints, all planning runs were given a \emph{1 minute timeout} and 4GB memory limit.
Feature pruning is not supported for the \iwl{} and \niwl{} algorithms.
In summary, we have
$
2 \times 8 \times 2 \times 2 \times 2 \times 6 +
2 \times 8 \times 1 \times 2 \times 2 \times 6 = 1152
$
different model configurations over 10 domains, and 900 problems.
Thus, we have up to $1152 \times 10 = 11520$ learned models for up to $1152 \times 900 = 1036800$ planning runs.

\begin{figure}[t]
  \centering
  \includegraphics{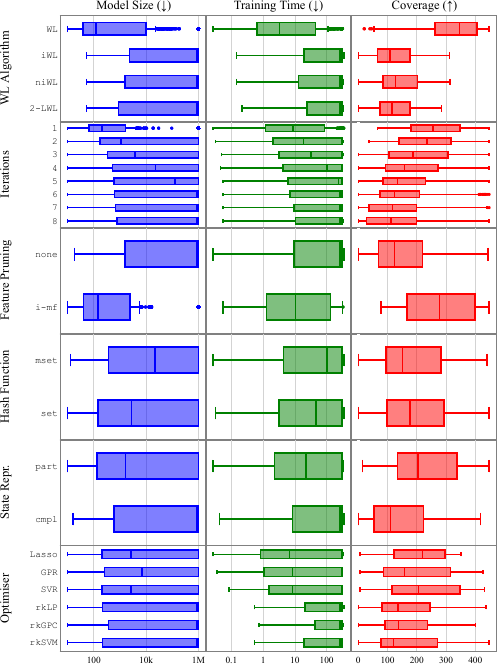}
  \caption{
    Boxplots of
    WLF counts (left, log-scale),
    training time in seconds (middle, log-scale), and
    coverage (right) of WLF models conditioned on hyperparameter options.
    Arrows indicate whether lower ($\downarrow$) or higher ($\uparrow$) values are better.
    If no model is learned under the resource constraints for a hyperparameter set, its model size and training time value is set to 1e6 and 300, respectively.
  }
  \label{fig:n_colours}
  \label{fig:train_time}
  \label{fig:coverage}
\end{figure}

\paragraph{Which hyperparameters provide the smallest and fastest training models?}
The first two rows of \Cref{fig:n_colours} illustrate distributions of model size and training time of various models conditioned on different hyperparameter configurations.
For all internal hyperparameters (top half of \Cref{tab:hyperparameters}), the ranking of model size and train time is correlated.
The most efficient configuration from each hyperparameter choice is (1) \wl{}, (2) \texttt{1}, (3) \collapselayeryf{}, (4) \texttt{set}, (5) \ppartial, and (6) \lasso{}.
These results are not too surprising as discussed in their introductions.
One note however is that ranking methods generally take longer to train due to the introduction of pairwise comparisons between data compared to regression methods (\svr{}, \gpr{}, \lasso{}).

\paragraph{Which hyperparameters provide the overall best planning performance?}
The third column of \Cref{fig:n_colours} illustrates the distribution of planning coverage of various models, while \Cref{tab:coverage} displays the scores of the best performing configuration, both conditioned on different hyperparameter configurations.
The hyperparameters in each group except Optimiser with the best median performance also match those with the best maximal performance (\domainFont{M$\Sigma$} column in \Cref{tab:coverage}).
This exception for the Optimiser group is because the best (\ranksvm) models are more expensive to train and thus its median performance is degraded by configurations using other expensive hyperparameters.
Regardless, we decide that choosing hyperparameters based on the best overall, maximal performance is the best starting point as median performance is influenced by suboptimal hyperparameter choices.
To summarise from the data, the best overall choice of hyperparameters for WLFs are
(1) $\wl$, (2) $\texttt{1}$, (3) $\collapselayeryf$, (4) $\sset$, (5) $\ppartial$, (6) $\ranksvm$.
We note that there are exceptions on a per domain basis.
This is the case for Feature Pruning and Iterations ($\texttt{none}$ and $\texttt{2}$ are better) for Blocksworld, State Representation ($\texttt{cmpl}$ is better) for Childsnack, and Optimiser ($\svr$ is better) for Satellite.
\Cref{app:top,app:versus} provide additional details and metrics concerning top performing configurations and how they perform compared to the original WLF configuration in \cite{chen.etal.icaps2024}.

\paragraph{Is there any correlation between training and planning metrics?}
Our final question aims to answer whether training metrics give us a sense of planning performance.
In order to answer this question, we analyse the Pearson's correlation coefficient $\rho$ between planning coverage and optimiser evaluation function (Eval), training time (Time), and model size (Size) summarised in \Cref{tab:correlation}, and more fine grained results per domain provided in \Cref{app:corr}.
Unfortunately, we do not find any statistically significant ($p < 0.05$), strong correlation ($\abs{\rho} \geq 0.5$) between any training metric and planning performance for any optimiser.
This is not surprising for classical and statistical machine learning methods as in this study, which are bound to the bias-variance tradeoff at least when analysing training evaluation metrics.
Related and concurrent work study model selection using validation sets for policies~\cite{gros.etal.icaps2025}.

\begin{table}[t]
  \centering
  \setlength{\tabcolsep}{1pt}
  \newcommand{\groupFontSize}{\scriptsize}
  \newcommand{\domainFontSize}{\scriptsize}
  \newcommand{\splitmidrule}{\midrule}
  \begin{threeparttable}
    \begin{tabularx}{\columnwidth}{l *{12}{Y}}
    \toprule
    & \domainFontSize \domainFont{$\Sigma$M}
    & \domainFontSize \domainFont{M$\Sigma$}
    & \domainFontSize \bl{} 
    & \domainFontSize \ch{} 
    & \domainFontSize \fe{} 
    & \domainFontSize \fl{} 
    & \domainFontSize \mi{} 
    & \domainFontSize \ro{} 
    & \domainFontSize \sa{} 
    & \domainFontSize \so{} 
    & \domainFontSize \sp{} 
    & \domainFontSize \tr{}
    \\
    \midrule
    \groupFontSize WL Algorithm$^\ddagger$ \\
    \splitmidrule

\wl & \first{513} & \first{447} & \first{66} & \first{47} & \first{66} & \first{3} & \first{88} & \first{45} & \first{51} & \first{33} & \first{64} & \first{50} \\ 
\iwl & \third{326} & \third{311} & 24 & \third{30} & \second{50} & \first{3} & \second{58} & \third{31} & \second{31} & \third{30} & \second{38} & \third{31} \\ 
\niwl & \second{336} & \second{314} & \third{28} & \second{31} & \second{50} & \second{2} & \second{58} & \second{32} & \second{31} & \second{31} & \second{38} & \second{35} \\ 
\lwl & 321 & 285 & \second{31} & \third{30} & \third{48} & \second{2} & \third{54} & 30 & \third{30} & 28 & \third{37} & \third{31} \\

    \midrule
    \groupFontSize Iterations$^{\ddagger,*}$ \\
    \splitmidrule

\texttt{1} & \first{497} & \first{447} & 50 & \first{47} & \first{66} & \first{3} & \first{88} & \first{45} & \first{51} & \first{33} & \first{64} & \first{50} \\ 
\texttt{2} & \second{474} & \second{421} & \first{66} & \first{47} & \second{65} & \first{3} & \second{83} & \second{37} & 35 & \first{33} & \second{63} & \second{42} \\ 
\texttt{3} & \third{464} & 404 & \first{66} & \second{45} & \second{65} & \first{3} & \third{81} & \third{34} & 36 & \first{33} & \second{63} & \third{38} \\ 
\texttt{4} & 448 & \third{410} & \third{57} & 42 & \third{64} & \second{2} & 79 & \third{34} & 37 & \first{33} & \second{63} & 37 \\ 
\texttt{5} & 453 & 400 & \second{58} & \third{44} & 63 & \second{2} & 78 & \third{34} & \second{42} & \first{33} & \third{62} & 37 \\ 
\texttt{6} & 439 & 396 & 55 & 40 & 63 & \second{2} & 77 & 33 & \third{40} & \second{32} & \third{62} & 35 \\ 
\texttt{7} & 434 & 402 & 55 & 39 & 62 & \second{2} & 75 & 32 & \third{40} & \second{32} & \third{62} & 35 \\ 
\texttt{8} & 432 & 400 & 52 & 41 & 62 & \third{1} & 74 & 32 & \third{40} & \second{32} & \third{62} & 36 \\

    \midrule
    \groupFontSize Feature Pruning$^{\dagger}$ \\
    \splitmidrule

\none & \first{504} & 444 & \first{66} & \first{47} & \first{66} & \first{3} & 87 & 42 & 46 & \first{33} & \first{64} & \first{50} \\ 
\collapselayeryf & 492 & \first{447} & 50 & 46 & \first{66} & \first{3} & \first{88} & \first{45} & \first{51} & \first{33} & 63 & 47 \\

    \midrule
    \groupFontSize Hash Function$^{\dagger}$ \\
    \splitmidrule

\multiset & 502 & 440 & \first{66} & 46 & 65 & \first{3} & \first{88} & 43 & 50 & \first{33} & \first{64} & 44 \\ 
\sset & \first{512} & \first{447} & \first{66} & \first{47} & \first{66} & \first{3} & 87 & \first{45} & \first{51} & \first{33} & \first{64} & \first{50} \\

    \midrule
    \groupFontSize State Repr.$^{\dagger}$ \\
    \splitmidrule

\ppartial & \first{499} & \first{447} & \first{66} & 33 & \first{66} & \first{3} & \first{88} & \first{45} & \first{51} & \first{33} & \first{64} & \first{50} \\ 
\complete & 444 & 418 & 62 & \first{47} & 63 & 2 & 64 & 30 & 46 & 32 & 61 & 37 \\

    \midrule
    \groupFontSize Optimiser$^\ddagger$ \\
    \splitmidrule

\lasso & 388 & 351 & 34 & 30 & \second{65} & \third{1} & \first{88} & 31 & 31 & \first{33} & \third{36} & 39 \\ 
\gpr & 447 & 427 & \third{55} & 22 & \second{65} & \second{2} & \second{87} & \third{41} & 43 & \first{33} & \second{63} & 36 \\ 
\svr & 471 & \third{431} & \second{58} & 30 & \second{65} & \second{2} & \second{87} & \second{42} & \first{51} & \second{32} & \second{63} & \third{41} \\ 
\ranklp & \second{498} & \second{437} & \first{66} & \first{47} & \first{66} & \first{3} & \first{88} & 40 & \third{44} & \first{33} & \first{64} & \second{47} \\ 
\rankgpc & \third{480} & 401 & \first{66} & \third{45} & \second{65} & \first{3} & \third{86} & \third{41} & 31 & \first{33} & \second{63} & \second{47} \\ 
\ranksvm & \first{506} & \first{447} & \first{66} & \second{46} & \first{66} & \first{3} & \second{87} & \first{45} & \second{46} & \first{33} & \first{64} & \first{50} \\

\midrule\midrule
\groupFontSize
Orig. Conf. \cite{chen.etal.icaps2024} 
& -- & 398 & 55 & 15 & 63 & 1 & 79 & 34 & 33 & 32 & 60 & 26 \\

    \bottomrule
  \end{tabularx}
  
    \begin{tablenotes}
    \item $^\dagger$ The top value in the column is highlighted.
    \item $^\ddagger$ The top 3 values per column are highlighted.
    \item $^*$ The Feature Pruning option \imf{} sometimes reduces the number of iterations. These occurrences are omitted from the Iterations rows.
    \end{tablenotes}
  \end{threeparttable}

  \caption{
    Best performance ($\uparrow$) conditioned on hyperparameter option and planning domain.
    The \domainFont{$\Sigma$M} column sums over the domain values, while the \domainFont{M$\Sigma$} column takes the maximum total coverage conditioned on the row feature.
    The training timeout is \emph{5 minutes} and memory limit is 8GB.
    The planning timeout is \emph{60 seconds} and memory limit is 4GB.
    The final row (Orig. Conf.) corresponds to hyperparameters (\wl{}, \texttt{4}, \texttt{none}, \texttt{mset}, \texttt{part}, \texttt{GPR}) in the original WLF paper~\cite{chen.etal.icaps2024} using the resource limits in this study.
  }
  \label{tab:coverage}
\end{table}

\newcommand{\statsize}{\tiny}
\newcommand{\notstat}[1]{\cellcolor{gray!25}{\statsize{#1}}}
\newcommand{\sigstat}[1]{\cellcolor{white}{\statsize{#1}}}
\begin{table}[t]
  \tabcolsep 1pt
  \begin{tabularx}{\columnwidth}{l *{6}{Y}}
  \toprule
  & \lasso{}
  & \gpr{}
  & \svr{}
  & \ranklp{}
  & \rankgpc{}
  & \ranksvm{}         \\
  \midrule
Eval & \notstat{[\texttt{-.10}, \texttt{ .00}]} & \notstat{[\texttt{-.05}, \texttt{ .06}]} & \notstat{[\texttt{-.10}, \texttt{ .01}]} & \sigstat{[\texttt{-.32}, \texttt{-.21}]} & \sigstat{[\texttt{ .12}, \texttt{ .25}]} & \sigstat{[\texttt{ .07}, \texttt{ .19}]}\\
Time & \notstat{[\texttt{-.09}, \texttt{ .01}]} & \notstat{[\texttt{-.07}, \texttt{ .04}]} & \notstat{[\texttt{-.07}, \texttt{ .04}]} & \sigstat{[\texttt{-.13}, \texttt{-.00}]} & \sigstat{[\texttt{-.15}, \texttt{-.02}]} & \sigstat{[\texttt{-.15}, \texttt{-.02}]}\\
Size & \sigstat{[\texttt{-.22}, \texttt{-.12}]} & \sigstat{[\texttt{-.20}, \texttt{-.09}]} & \sigstat{[\texttt{-.23}, \texttt{-.12}]} & \sigstat{[\texttt{-.30}, \texttt{-.18}]} & \sigstat{[\texttt{-.14}, \texttt{-.01}]} & \sigstat{[\texttt{-.38}, \texttt{-.27}]}\\
  \bottomrule
\end{tabularx}

  \caption{
    Confidence intervals of Pearson's correlation coefficient between training metrics (rows) and planning coverage of optimisers (columns).
    Statistically insignificant cells ($p \geq 0.05$) are indicated in gray.
  }
  \label{tab:correlation}
\end{table}

\section{Conclusion}
In this paper, we have introduced several new hyperparameters associated with Weisfeiler-Leman Features (WLFs) in the context of learning to plan.
We focus on the task of learning heuristics in this paper, although WLFs are agnostic to the downstream planning task.
We classified the WLF hyperparameters and performed an experimental evaluation with a 1,000,000 sample size in order to understand the effects of various hyperparameters.
Our evaluation aimed to answer three core questions related to the effect of hyperparameters on (1) training, (2) planning, and (3) the correlation between training and planning metrics in the context of learning to plan.
Indeed, from our analysis we have identified a best set of hyperparameters with almost consistent performance across different planning domains.

\paragraph{Acknowledgements}
The author thanks the anonymous reviewers for their comments that helped improve this paper.

\bibliography{ecai-25-hyperparameters.bib}

\appendix
\onecolumn
\tableofcontents
\clearpage

\section{Additional Benchmark Details}\label{app:benchmarks}
\Cref{tab:benchmarks} illustrates the number of states and label range in the training data, and the number of objects in the training and testing problems.

\begin{table}[h!]
  \begin{tabularx}{\textwidth}{X Y Y Y}
    \toprule
    Domain & Number of Labelled States (concerns \lasso{}, \gpr{}, \svr{}) & Number of States and Siblings (concerns \ranklp{}, \rankgpc{}, \ranksvm{}) & Maximum Label \\
    \midrule
    Blocksworld &
    1292
    &
    7703
    &
    56
    \\
    Childsnack &
    473
    &
    10334
    &
    18
    \\
    Ferry &
    1368
    &
    13719
    &
    45
    \\
    Floortile &
    3187
    &
    32549
    &
    83
    \\
    Miconic &
    1630
    &
    24089
    &
    35
    \\
    Rovers &
    1428
    &
    21036
    &
    39
    \\
    Satellite &
    1158
    &
    65312
    &
    37
    \\
    Sokoban &
    2422
    &
    10406
    &
    67
    \\
    Spanner &
    1204
    &
    4576
    &
    21
    \\
    Transport &
    735
    &
    15013
    &
    29
    \\
    \bottomrule
  \end{tabularx}
  \newcommand{\splitter}{
  }
  \vspace{6pt}
  \begin{tabularx}{\textwidth}{X X Y Y Y Y}
    \toprule
    &
    &
    & {Train} &
    & {Test} \\
    {Domain} & {Object Types} & Min & Max & Min & Max \\
    \midrule
    Blocksworld
    & $\Sigma$ & 2 & 29 & 5 & 488 \\
    \splitter
    & blocks & 2 & 29 & 5 & 488 \\
    \midrule
    Childsnack
    & $\Sigma$ & 6 & 51 & 20 & 1326 \\
    \splitter
    & bread-portion & 1 & 10 & 4 & 292 \\
    & child & 1 & 10 & 4 & 292 \\
    & content-portion & 1 & 10 & 4 & 292 \\
    & place & 1 & 3 & 3 & 3 \\
    & sandwich & 1 & 15 & 4 & 437 \\
    & tray & 1 & 3 & 1 & 10 \\
    \midrule
    Ferry
    & $\Sigma$ & 3 & 35 & 7 & 1461 \\
    \splitter
    & car & 1 & 20 & 2 & 974 \\
    & location & 2 & 15 & 5 & 487 \\
    \midrule
    Floortile
    & $\Sigma$ & 5 & 45 & 15 & 1022 \\
    \splitter
    & color & 2 & 2 & 2 & 2 \\
    & robot & 1 & 3 & 1 & 34 \\
    & tile & 2 & 40 & 12 & 986 \\
    \midrule
    Miconic
    & $\Sigma$ & 3 & 30 & 5 & 681 \\
    \splitter
    & floor & 2 & 20 & 4 & 196 \\
    & passenger & 1 & 10 & 1 & 485 \\
    \midrule
    Rovers
    & $\Sigma$ & 10 & 36 & 12 & 596 \\
    \splitter
    & camera & 1 & 4 & 1 & 99 \\
    & lander & 1 & 1 & 1 & 1 \\
    & mode & 3 & 3 & 3 & 3 \\
    & objective & 1 & 10 & 1 & 236 \\
    & rover & 1 & 4 & 1 & 30 \\
    & store & 1 & 4 & 1 & 30 \\
    & waypoint & 2 & 10 & 4 & 197 \\
    \midrule
    Satellite
    & $\Sigma$ & 5 & 43 & 11 & 402 \\
    \splitter
    & direction & 2 & 10 & 4 & 98 \\
    & instrument & 1 & 20 & 3 & 195 \\
    & mode & 1 & 3 & 1 & 10 \\
    & satellite & 1 & 10 & 3 & 99 \\
    \midrule
    Sokoban
    & $\Sigma$ & 50 & 173 & 65 & 9880 \\
    \splitter
    & box & 1 & 4 & 1 & 79 \\
    & location & 49 & 169 & 64 & 9801 \\
    \midrule
    Spanner
    & $\Sigma$ & 6 & 28 & 9 & 833 \\
    \splitter
    & location & 3 & 12 & 6 & 101 \\
    & man & 1 & 1 & 1 & 1 \\
    & nut & 1 & 5 & 1 & 244 \\
    & spanner & 1 & 10 & 1 & 487 \\
    \midrule
    Transport
    & $\Sigma$ & 6 & 45 & 12 & 354 \\
    \splitter
    & location & 2 & 17 & 5 & 99 \\
    & package & 1 & 17 & 1 & 194 \\
    & size & 2 & 4 & 3 & 11 \\
    & vehicle & 1 & 7 & 3 & 50 \\
    \bottomrule
  \end{tabularx}
  \caption{\emph{Top:} labelled training data details. \emph{Bottom:} distributions of object sizes by object type in training and testing splits over tested domains.}
  \label{tab:benchmarks}
\end{table}

\clearpage
\section{Best Performing Configurations}\label{app:top}
\Cref{tab:top} lists the top performing configurations and the configuration in the original WLF for planning paper~\cite{chen.etal.icaps2024} with respect to the International Planning Competition (IPC) agile track score.
Specifically, a configuration gets a score for a problem given by $1$ if the problem is solved in less than a second, $1 - \frac{\log(t)}{\log(T)}$ if the problem is solved in $t$ seconds, and $T=60$ is the planning timeout, and $0$ otherwise, i.e. if the problem is not solved within the timeout or memory limit.
The total agile score for a configuration is given by the sum of the score over all problems.

\begin{table}
  \centering
  
    \begin{tabularx}{\textwidth}{*{6}{X} *{2}{Y}}
        \toprule
        WL Algorithm & 
        Iterations & 
        Feature Pruning & 
        Hash Function & 
        State Repr. &
        Optimiser &
        Agile Score ($\uparrow$) &
        Coverage ($\uparrow$) \\
        \cmidrule{1-6}
        \cmidrule(l){7-8}
        \wl & \texttt{1} & \collapselayeryf & \sset & \ppartial & \ranksvm & 386.03 & 447\\
\wl & \texttt{1} & \collapselayeryf & \multiset & \ppartial & \ranksvm & 383.94 & 440\\
\wl & \texttt{1} & \none & \sset & \ppartial & \ranksvm & 375.88 & 444\\
\wl & \texttt{1} & \none & \sset & \ppartial & \ranklp & 374.72 & 437\\
\wl & \texttt{1} & \collapselayeryf & \sset & \ppartial & \ranklp & 374.59 & 435\\
\wl & \texttt{1} & \collapselayeryf & \multiset & \ppartial & \ranklp & 371.50 & 424\\
\wl & \texttt{1} & \none & \sset & \ppartial & \svr & 370.63 & 430\\
\wl & \texttt{1} & \collapselayeryf & \sset & \ppartial & \svr & 370.52 & 431\\
\wl & \texttt{1} & \none & \multiset & \ppartial & \ranksvm & 370.24 & 426\\
\wl & \texttt{1} & \none & \multiset & \ppartial & \ranklp & 368.79 & 426\\
\wl & \texttt{2} & \none & \sset & \ppartial & \ranksvm & 367.02 & 420\\
\wl & \texttt{2} & \none & \sset & \ppartial & \ranklp & 365.86 & 421\\
\wl & \texttt{1} & \collapselayeryf & \multiset & \ppartial & \svr & 365.13 & 412\\
\wl & \texttt{1} & \collapselayeryf & \sset & \ppartial & \gpr & 362.69 & 427\\
\wl & \texttt{1} & \none & \sset & \ppartial & \gpr & 362.12 & 416\\
\wl & \texttt{1} & \none & \multiset & \ppartial & \svr & 357.64 & 420\\
\wl & \texttt{1} & \collapselayeryf & \multiset & \ppartial & \gpr & 353.51 & 413\\
\wl & \texttt{1} & \none & \multiset & \ppartial & \gpr & 350.00 & 414\\
\wl & \texttt{1} & \collapselayeryf & \sset & \ppartial & \rankgpc & 349.56 & 398\\
\wl & \texttt{3} & \none & \multiset & \ppartial & \svr & 348.86 & 404\\
\midrule
\wl & \texttt{4} & \none & \multiset & \ppartial & \gpr & 341.54 & 398\\
\bottomrule
    \end{tabularx}
  \caption{Top 20 performing configurations ranked by IPC agile score and the configuration from the original WLF paper.}
  \label{tab:top}
\end{table}

\section{Best Performing Configuration vs. Original Configuration}\label{app:versus}
\begin{figure}
  \centering
  \includegraphics{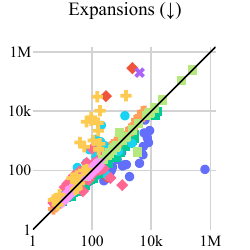}
  \includegraphics{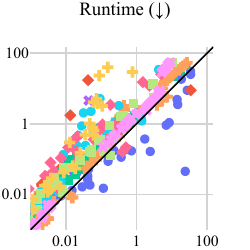}
  \includegraphics{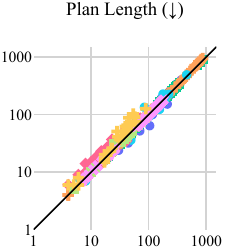}
  \includegraphics{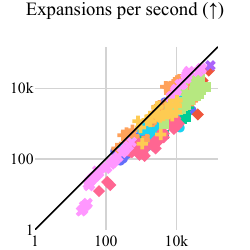}
  \caption{
    Various metrics for (1) the best WLF configuration vs. (2) the configuration from the original WLF paper.
    Note the log scales.
    The arrow directions indicate whether higher or lower values are better for a metric.
    Upper left points favour (1) for $\uparrow$ metrics, and lower right points value (1) for $\downarrow$ metrics.
  }
  \label{fig:versus}
\end{figure}

\begin{figure}
  \centering
  \includegraphics{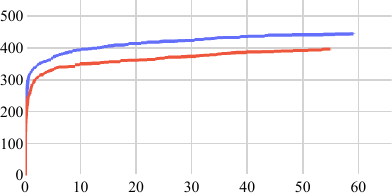}
  \caption{
    Cumulative coverage ($\uparrow$) over time for the best WLF configuration (blue) and the configuration from the original WLF paper (red).
  }
  \label{fig:cumm}
\end{figure}

\Cref{fig:versus,fig:cumm} compare the best performing configuration against original WLF configuration~\cite{chen.etal.icaps2024} over various metrics.

\clearpage
\section{Statistical Correlation by Domain}\label{app:corr}
\begin{table}
  \centering
  \renewcommand{\statsize}{}
  \begin{tabularx}{\columnwidth}{l *{6}{Y}}
  \toprule
  & \lasso{}
  & \gpr{}
  & \svr{}
  & \ranklp{}
  & \rankgpc{}
  & \ranksvm{}         \\

\midrule
Blocksworld\\
\midrule
Eval & \sigstat{[\texttt{-.80}, \texttt{-.63}]} & \notstat{[\texttt{-.07}, \texttt{ .28}]} & \sigstat{[\texttt{-.39}, \texttt{-.06}]} & \notstat{[\texttt{-.32}, \texttt{ .11}]} & \notstat{[\texttt{-.31}, \texttt{ .08}]} & \notstat{[\texttt{-.30}, \texttt{ .10}]}\\
Time & \notstat{[\texttt{-.07}, \texttt{ .27}]} & \sigstat{[\texttt{ .14}, \texttt{ .46}]} & \sigstat{[\texttt{ .24}, \texttt{ .53}]} & \notstat{[\texttt{-.38}, \texttt{ .04}]} & \sigstat{[\texttt{-.49}, \texttt{-.13}]} & \sigstat{[\texttt{-.47}, \texttt{-.10}]}\\
Size & \sigstat{[\texttt{-.53}, \texttt{-.24}]} & \notstat{[\texttt{-.22}, \texttt{ .13}]} & \sigstat{[\texttt{-.40}, \texttt{-.07}]} & \sigstat{[\texttt{-.68}, \texttt{-.38}]} & \sigstat{[\texttt{-.53}, \texttt{-.19}]} & \sigstat{[\texttt{-.68}, \texttt{-.39}]}\\
\midrule
Childsnack\\
\midrule
Eval & \sigstat{[\texttt{ .92}, \texttt{ .95}]} & \sigstat{[\texttt{-.58}, \texttt{-.36}]} & \notstat{[\texttt{-.17}, \texttt{ .11}]} & \sigstat{[\texttt{ .46}, \texttt{ .66}]} & \notstat{[\texttt{-.20}, \texttt{ .10}]} & \sigstat{[\texttt{-.46}, \texttt{-.19}]}\\
Time & \sigstat{[\texttt{ .00}, \texttt{ .28}]} & \sigstat{[\texttt{-.30}, \texttt{-.03}]} & \notstat{[\texttt{-.22}, \texttt{ .06}]} & \notstat{[\texttt{-.25}, \texttt{ .05}]} & \sigstat{[\texttt{-.46}, \texttt{-.19}]} & \sigstat{[\texttt{-.31}, \texttt{-.02}]}\\
Size & \sigstat{[\texttt{ .09}, \texttt{ .36}]} & \sigstat{[\texttt{-.46}, \texttt{-.21}]} & \sigstat{[\texttt{-.41}, \texttt{-.14}]} & \sigstat{[\texttt{-.34}, \texttt{-.05}]} & \sigstat{[\texttt{-.59}, \texttt{-.35}]} & \sigstat{[\texttt{-.64}, \texttt{-.43}]}\\
\midrule
Ferry\\
\midrule
Eval & \sigstat{[\texttt{-.84}, \texttt{-.71}]} & \sigstat{[\texttt{-.44}, \texttt{-.15}]} & \sigstat{[\texttt{-.38}, \texttt{-.09}]} & \sigstat{[\texttt{ .33}, \texttt{ .61}]} & \notstat{[\texttt{-.09}, \texttt{ .28}]} & \notstat{[\texttt{-.16}, \texttt{ .20}]}\\
Time & \sigstat{[\texttt{-.37}, \texttt{-.07}]} & \sigstat{[\texttt{-.32}, \texttt{-.01}]} & \sigstat{[\texttt{-.33}, \texttt{-.03}]} & \sigstat{[\texttt{-.40}, \texttt{-.05}]} & \sigstat{[\texttt{-.45}, \texttt{-.11}]} & \sigstat{[\texttt{-.42}, \texttt{-.08}]}\\
Size & \sigstat{[\texttt{-.41}, \texttt{-.11}]} & \sigstat{[\texttt{-.38}, \texttt{-.09}]} & \sigstat{[\texttt{-.40}, \texttt{-.10}]} & \sigstat{[\texttt{-.55}, \texttt{-.25}]} & \notstat{[\texttt{-.30}, \texttt{ .06}]} & \sigstat{[\texttt{-.83}, \texttt{-.67}]}\\
\midrule
Floortile\\
\midrule
Eval & \sigstat{[\texttt{ .09}, \texttt{ .43}]} & \notstat{[\texttt{-.18}, \texttt{ .26}]} & \sigstat{[\texttt{-.41}, \texttt{-.07}]} & \notstat{[\texttt{-.25}, \texttt{ .27}]} & \notstat{[\texttt{-.01}, \texttt{ .64}]} & \sigstat{[\texttt{-.63}, \texttt{-.18}]}\\
Time & \notstat{[\texttt{-.06}, \texttt{ .30}]} & \notstat{[\texttt{-.42}, \texttt{ .00}]} & \notstat{[\texttt{-.18}, \texttt{ .18}]} & \notstat{[\texttt{-.33}, \texttt{ .18}]} & \sigstat{[\texttt{-.68}, \texttt{-.05}]} & \notstat{[\texttt{-.27}, \texttt{ .27}]}\\
Size & \sigstat{[\texttt{-.37}, \texttt{-.02}]} & \notstat{[\texttt{-.28}, \texttt{ .16}]} & \sigstat{[\texttt{-.61}, \texttt{-.33}]} & \sigstat{[\texttt{-.53}, \texttt{-.06}]} & \sigstat{[\texttt{-.89}, \texttt{-.56}]} & \notstat{[\texttt{-.36}, \texttt{ .18}]}\\
\midrule
Miconic\\
\midrule
Eval & \sigstat{[\texttt{-.63}, \texttt{-.39}]} & \notstat{[\texttt{-.18}, \texttt{ .20}]} & \sigstat{[\texttt{-.36}, \texttt{-.04}]} & \notstat{[\texttt{-.11}, \texttt{ .26}]} & \sigstat{[\texttt{ .06}, \texttt{ .42}]} & \notstat{[\texttt{-.13}, \texttt{ .26}]}\\
Time & \notstat{[\texttt{-.25}, \texttt{ .09}]} & \notstat{[\texttt{-.30}, \texttt{ .07}]} & \notstat{[\texttt{-.26}, \texttt{ .07}]} & \sigstat{[\texttt{-.55}, \texttt{-.24}]} & \sigstat{[\texttt{-.57}, \texttt{-.25}]} & \sigstat{[\texttt{-.59}, \texttt{-.28}]}\\
Size & \sigstat{[\texttt{-.54}, \texttt{-.26}]} & \sigstat{[\texttt{-.49}, \texttt{-.14}]} & \sigstat{[\texttt{-.43}, \texttt{-.12}]} & \sigstat{[\texttt{-.63}, \texttt{-.35}]} & \sigstat{[\texttt{-.61}, \texttt{-.31}]} & \sigstat{[\texttt{-.83}, \texttt{-.66}]}\\
\midrule
Rovers\\
\midrule
Eval & \sigstat{[\texttt{ .17}, \texttt{ .53}]} & \notstat{[\texttt{-.03}, \texttt{ .37}]} & \sigstat{[\texttt{-.47}, \texttt{-.10}]} & \sigstat{[\texttt{-.78}, \texttt{-.41}]} & \sigstat{[\texttt{ .47}, \texttt{ .87}]} & \notstat{[\texttt{-.14}, \texttt{ .42}]}\\
Time & \notstat{[\texttt{-.18}, \texttt{ .22}]} & \notstat{[\texttt{-.17}, \texttt{ .25}]} & \notstat{[\texttt{-.32}, \texttt{ .08}]} & \sigstat{[\texttt{-.73}, \texttt{-.30}]} & \sigstat{[\texttt{-.93}, \texttt{-.71}]} & \sigstat{[\texttt{-.62}, \texttt{-.12}]}\\
Size & \sigstat{[\texttt{-.57}, \texttt{-.23}]} & \sigstat{[\texttt{-.63}, \texttt{-.30}]} & \sigstat{[\texttt{-.68}, \texttt{-.40}]} & \sigstat{[\texttt{-.89}, \texttt{-.65}]} & \sigstat{[\texttt{-.69}, \texttt{-.06}]} & \sigstat{[\texttt{-.70}, \texttt{-.26}]}\\
\midrule
Satellite\\
\midrule
Eval & \sigstat{[\texttt{ .76}, \texttt{ .88}]} & \notstat{[\texttt{-.19}, \texttt{ .19}]} & \notstat{[\texttt{-.11}, \texttt{ .25}]} & \sigstat{[\texttt{ .12}, \texttt{ .65}]} & \sigstat{[\texttt{ .02}, \texttt{ 1.00}]} & \sigstat{[\texttt{ .20}, \texttt{ .69}]}\\
Time & \notstat{[\texttt{-.29}, \texttt{ .07}]} & \notstat{[\texttt{-.09}, \texttt{ .28}]} & \notstat{[\texttt{-.23}, \texttt{ .13}]} & \notstat{[\texttt{-.56}, \texttt{ .03}]} & \notstat{[\texttt{-.97}, \texttt{ .94}]} & \sigstat{[\texttt{-.81}, \texttt{-.44}]}\\
Size & \sigstat{[\texttt{ .16}, \texttt{ .48}]} & \sigstat{[\texttt{-.48}, \texttt{-.14}]} & \sigstat{[\texttt{-.44}, \texttt{-.10}]} & \sigstat{[\texttt{-.61}, \texttt{-.05}]} & \sigstat{[\texttt{-1.00}, \texttt{ .03}]} & \sigstat{[\texttt{-.76}, \texttt{-.33}]}\\
\midrule
Sokoban\\
\midrule
Eval & \sigstat{[\texttt{ .53}, \texttt{ .74}]} & \notstat{[\texttt{-.03}, \texttt{ .41}]} & \sigstat{[\texttt{-.65}, \texttt{-.39}]} & \notstat{[\texttt{-.27}, \texttt{ .16}]} & \sigstat{[\texttt{-.42}, \texttt{-.01}]} & \notstat{[\texttt{-.09}, \texttt{ .39}]}\\
Time & \sigstat{[\texttt{-.40}, \texttt{-.07}]} & \notstat{[\texttt{-.32}, \texttt{ .13}]} & \notstat{[\texttt{-.15}, \texttt{ .20}]} & \sigstat{[\texttt{-.48}, \texttt{-.09}]} & \sigstat{[\texttt{-.59}, \texttt{-.24}]} & \notstat{[\texttt{-.23}, \texttt{ .26}]}\\
Size & \sigstat{[\texttt{-.80}, \texttt{-.64}]} & \sigstat{[\texttt{-.72}, \texttt{-.42}]} & \sigstat{[\texttt{-.82}, \texttt{-.66}]} & \sigstat{[\texttt{-.69}, \texttt{-.40}]} & \sigstat{[\texttt{-.91}, \texttt{-.79}]} & \sigstat{[\texttt{-.96}, \texttt{-.89}]}\\
\midrule
Spanner\\
\midrule
Eval & \sigstat{[\texttt{ .46}, \texttt{ .66}]} & \notstat{[\texttt{-.17}, \texttt{ .14}]} & \sigstat{[\texttt{-.37}, \texttt{-.09}]} & \sigstat{[\texttt{ .04}, \texttt{ .33}]} & \sigstat{[\texttt{ .06}, \texttt{ .35}]} & \sigstat{[\texttt{ .03}, \texttt{ .33}]}\\
Time & \sigstat{[\texttt{-.41}, \texttt{-.13}]} & \sigstat{[\texttt{-.33}, \texttt{-.03}]} & \notstat{[\texttt{-.28}, \texttt{ .01}]} & \sigstat{[\texttt{-.33}, \texttt{-.04}]} & \notstat{[\texttt{-.28}, \texttt{ .02}]} & \sigstat{[\texttt{-.32}, \texttt{-.03}]}\\
Size & \notstat{[\texttt{-.27}, \texttt{ .02}]} & \sigstat{[\texttt{-.40}, \texttt{-.11}]} & \sigstat{[\texttt{-.44}, \texttt{-.17}]} & \sigstat{[\texttt{-.50}, \texttt{-.24}]} & \sigstat{[\texttt{-.55}, \texttt{-.31}]} & \sigstat{[\texttt{-.59}, \texttt{-.35}]}\\
\midrule
Transport\\
\midrule
Eval & \sigstat{[\texttt{-.88}, \texttt{-.78}]} & \sigstat{[\texttt{ .16}, \texttt{ .47}]} & \notstat{[\texttt{-.14}, \texttt{ .20}]} & \sigstat{[\texttt{ .46}, \texttt{ .76}]} & \sigstat{[\texttt{ .17}, \texttt{ .64}]} & \notstat{[\texttt{-.34}, \texttt{ .08}]}\\
Time & \sigstat{[\texttt{-.41}, \texttt{-.08}]} & \notstat{[\texttt{-.03}, \texttt{ .31}]} & \notstat{[\texttt{-.12}, \texttt{ .22}]} & \sigstat{[\texttt{-.70}, \texttt{-.33}]} & \sigstat{[\texttt{-.73}, \texttt{-.33}]} & \sigstat{[\texttt{-.65}, \texttt{-.32}]}\\
Size & \sigstat{[\texttt{-.44}, \texttt{-.12}]} & \notstat{[\texttt{-.27}, \texttt{ .07}]} & \sigstat{[\texttt{-.37}, \texttt{-.05}]} & \sigstat{[\texttt{-.72}, \texttt{-.38}]} & \notstat{[\texttt{-.45}, \texttt{ .11}]} & \sigstat{[\texttt{-.84}, \texttt{-.65}]}\\

  \bottomrule
\end{tabularx}

  \caption{
    Pearson's correlation coefficient between training metrics (rows) and planning coverage of optimisers (columns) per domain.
    Statistically insignificant coefficients ($p \geq 0.05$) are indicated with italicised fonts and gray.
  }
\end{table}

\end{document}